\documentclass{article}

\usepackage{setspace}
\usepackage{amsmath,amssymb,amsthm}

\usepackage{graphicx}
\usepackage{float}
\usepackage{pdflscape}
\usepackage{longtable}
\usepackage{multirow}
\usepackage{adjustbox}
\usepackage[flushleft]{threeparttable}
\usepackage{pdfpages}
\usepackage{wrapfig}
\usepackage{caption}
\usepackage{subcaption}
\usepackage{tabularx}
\usepackage{multirow}

\usepackage{booktabs}
\usepackage{longtable}
\usepackage{array}

\newcolumntype{P}[1]{>{\raggedright\arraybackslash}p{#1}}

\graphicspath{{./}{figures/}}

\usepackage{xcolor}
\usepackage{soul}
\usepackage{seqsplit}
\usepackage{indentfirst}
\usepackage{enumitem}
\usepackage[
colorlinks=true,
urlcolor=blue,     
linkcolor=red,     
citecolor=blue    
]{hyperref}

\usepackage{algorithm}
\usepackage{algpseudocode}

\algrenewcommand\algorithmicrequire{\textbf{Input:}}
\algrenewcommand\algorithmicensure{\textbf{Output:}}

\usepackage{titlesec}
\titleformat{\section}[hang]{\bfseries\filright}{\thesection}{1em}{}
\titleformat{\subsection}[hang]{\bfseries}{\thesubsection}{1em}{}
\titleformat{\subsubsection}[hang]{\bfseries}{\thesubsubsection}{1em}{}

\usepackage[titletoc]{appendix}
\usepackage{url}
\usepackage{natbib}
\usepackage{authblk}   

\title{\textbf{A Physics-Informed Hybrid Forward-Inverse Machine Learning Framework for High-Dimensional Sparse Recipe Reconstruction}}

\author{Md. Muqtadir Fuad}
\author{Md. Istiak Adnan}
\author{Ridwan Al Aziz\thanks{Corresponding author. Email: \texttt{ridwanalaziz@ipe.buet.ac.bd}}}
\affil{Department of Industrial and Production Engineering,\\
	Bangladesh University of Engineering and Technology (BUET),\\
	Dhaka 1000, Bangladesh}
\date{}   

\begin{document}
	\maketitle
	
	\begin{abstract}
		High-dimensional sparse recipe formulation in process industries involves nonlinear association of various covariates, bounded by sparsity constraints and multi-criteria tolerances. Traditional physics-based classical methods (such as Kubelka-Munk) fail to generate sparse candidate recipes properly due to their simplified assumptions and industrial process variations. This study proposes a hybrid physics-informed uncertainty-aware framework that is structured into three modules: (a) Module A, a multi-target forward quality model; (b) Module B, a sparse inverse reconstructor; and (c) Module C, a feasibility-constrained ranking layer. Then it has been validated in an industrial textile dyeing context with recipe-grouped cross-validation under multi-illuminant quality assessment comprising process and production information, target reflectance, K/S values, and spectral descriptors. Histogram Gradient Boosting (HGB) with spectral-only features ($R_\lambda, K/S$) and Gradient Boosting Regressor (GBDT) with recipe-process features outperforms other configurations with out-of-Fold MAE of $0.1655$ and $0.1523$ in the forward quality model. For inverse reconstruction, a component-wise median blend achieved a recipe-vector MAE of $0.0295$. Exploratory conditional treatment effect (CATE) suggests an observational association that alkali adjustment reduces error for red and pale shades ($\text{CATE} = -0.196$). This framework provides a human-in-the-loop decision-support tool that minimizes costly laboratory trials across industrial formulation domains.
	\end{abstract}
	
	\vspace{2mm}
	
	
	
	\section{Introduction}
	\label{sec:intro}
	
	Recipe reconstruction in process manufacturing industries like chemical formulation, polymer compounding, and textile dyeing with large candidate spaces is often challenging \citep{venkatasubramanian_2019}. Specifically, industries have physical and economic constraints that dictate their choice within boundaries to construct the same or nearly the same recipe with a subset of ingredients \citep{cho_2026}. Besides, the historical production logs are not often preserved digitally in many industries \citep{tao_2018}. Moreover, the demands are also concentrated in narrow, frequently ordered color/product regions rather than uniformly covering the high-dimensional combinatorial ingredient space. For instance, during color recipe reconstruction in a dyeing lab, an industry is bound to use 3-4 colors at most for a certain formulation due to economic and policy constraints \citep{Samanta2024}. Additionally, recipes must pass specific quality assessment techniques and tolerance limits across various illumination and industrial conditions \citep{Minchew2013, ASTM_2025, Mukthy2021, AlZahrani2026}.
	
	Several computational and machine learning approaches have been explored to keep pace with these formulation and quality constraints \citep{Ingle2025}. They are categorized into three paradigms: (a) forward surrogate predictors, (b) inverse metaheuristic or neural estimators, and (c) retrieval and modular systems. However, these isolated approaches often suffer from operational limitations \citep{Ingle2025}. Researchers have explored different types of strategies, methods, and frameworks. Neural network, support vector regression, and other machine learning-based algorithms are used to predict color coordinates from known recipes for forward modeling \citep{vadood2022, li2022, Haji2023}. Another group studied diverse frameworks using metaheuristics like Genetic Algorithms (GA), Ant Colony Optimization (ACO), NSGA-II, and PSO for inverse prediction \citep{Chaouch2022, Chaouch2019, zhou2025}. Industries traditionally use historical retrieval systems \citep{Furferi2016, Li2025}. Besides, several researchers have considered this traditional industry-established system as a baseline for their framework \citep{Furferi2016, qin2021}. Some studies used modular neural networks or neuro-fuzzy systems for
	recipe and concentration prediction \citep{yang2018, shamsnateri2018}. Forward models can't synthesize a recipe itself without being wrapped in a huge combinatorial search subset. Besides, many of these works are limited to pre-selected 3-4 dyes \citep{zhang2021, chaouch2020}. But a real-world dyeing lab uses dozens of candidates at the same time. Apart from inverse regressors, they suffer from metameric mapping \citep{Mukthy2021, zhou2025, souissi2024color, souissi2024dyeing}. Thus, no standalone method can generate sparse 3–4 component recipes that meet multi-illuminant tolerances and provide reliable confidence intervals at the same time.
	
	In the context of such an industrial scenario, this study proposes an integrated closed-loop framework that orchestrates the historical retrieval system along with inverse recipe generation as well as forward quality screening. This framework can be divided into three basic modules. Module A, which predicts multi-illuminant color differences. It also adds an uncertainty quantification layer that assesses the reliability of the prediction. In module B, it reconstructs a recipe from a set of dye libraries that imitate practical industrial dyeing procedures while strictly following a constrained recipe generation policy. In module C, this framework pools similar or nearly similar historical recipes and uses a forward predictor module to assess the output quality. Then it ranks all feasible candidates against different quality parameters.
	
	The proposed methodology is formulated as a general framework for sparse recipe reconstruction and is validated the methodology on an industrial textile color matching problem in the proposed framework. Spectral and colorimetric systems serves as the industry standard. Factories use spectrophotometers and computer color-matching (CCM) softwares \citep{RahmanEtAl2024, Chaouch2024,sahin-2024}. A buyer provides either a fabric sample or a Pantone color code. A spectrophotometer scans the sample and records its spectral reflectance values. The CCM software analyses these values using stored information on fabric types, dyes, and chemicals and estimates the initial dye concentration. Laboratory technicians use this recipe to dye a small fabric sample known as a lab dip \citep{Gooby2020}. The lab dip is compared with the buyer’s standard. If the color difference is within the acceptable tolerance, the sample is sent to the buyer for approval. After approval, bulk production begins \citep{Abuhay2020}. Otherwise, the technician adjusts the recipe and repeats the test \citep{He2009}. This trial-and-error is the mathematical limitations of the conventional software. Most current systems rely on Kubelka-Munk (K-M) theory which assumes a linear relationship between dye concentration and fabric light absorption \citep{safi2023, safi-2021}. But, dyeing in the factory is a complicated, non-linear process. K-M assumptions cannot account for all dependencies such as water quality, temperature shifts, chemical additives, fiber inconsistencies etc, leading to a poor Right-First-Time (RFT) \citep{hu-2026}. But due to buyers' product-specific color requirements, factories requires to keep dyed samples within buyer-specified color-difference tolerances \citep{aatcc-2025,datacolor-tolerance}.
	
	The primary objective of this study is to develop a machine learning framework for textile dye recipe prediction that improves upon traditional linear modelling methods. This paper addresses the following research questions:
	
	\begin{enumerate}[label=\textbf{RQ\arabic*:}, leftmargin=*]
		
		\item How effectively can machine-learning models predict textile color quality using industrial recipe, process, and spectral data?
		\item How effectively can historical retrieval and inverse modelling produce useful dye-recipe candidates for known and unseen target shades?
		\item How robust and reliable are the resulting predictions, uncertainty estimates, and candidate recommendations under grouped validation?
		
	\end{enumerate}
	
	\section{Literature Review}
	\label{sec:review}
	In pre-1970s and even in some cases of the modern era, color recipe formulation relied on the expertise and practical knowledge of the colorists \citep{Chaouch2024, mcdonald1997}. But manual trial-and-error-based laboratory test limitations are attributed to production downtime and water waste. Most of these conventional approaches relied on the Kubelka-Munk (KM) theory. KM generates recipe formulations based on dye concentration and the fabrics' light absorption-to-scattering ratio (K/S) \citep{he2007,li2022, cutler1965}. Consequently, this formulation often fails to match the targeted shades under nuanced industrial settings in the first attempt. It leads to poor Right-First-Time (RFT) dye yield and creates industrial bottlenecks \citep{jeyaraj2014}.
	
	For the last twenty years, researchers have tried different machine-learning algorithms for the forward dye recipe formulation problem. In 2007 and later years, a group of researchers showed neural networks and their derivatives help to uncover the intricate non-linear behavior of dyeing \citep{li2007, almodarresi2013, shen2017, sun2023, sun2024}. Also, a backpropagation neural network with formula correction modules is being explored for dyeing performance improvement \citep{qin2025}. Researchers have used diversified sets of input variables such as fiber mixing proportions, dye concentrations, and operational parameters (temperature and dyeing time) to predict output spectral reflectance curves ($R_\lambda$), CIELAB color coordinates, and total color difference ($\Delta E$) \citep{sun2023, zhu2025, vadood2022, safi2023, jasper2025}.
	
	Researchers started using multi-criteria optimization with these complex algorithms in order to reconstruct recipes inversely. An example is the use of a least square support vector machine (LSSVM) and particle swarm optimization (PSO), which yielded lower errors \citep{yu2021, li2022}. Other researchers are also using such metaheuristic optimization algorithms along with hybrid neural networks to optimize feature space and solution space \citep{Chaouch2022,vadood2022, souissi2024color, souissi2024dyeing}. Similarly, Non-dominated Sorting Genetic Algorithm-II (NSGA-II) integrated with the Technique for Order of Preference by Similarity to Ideal Solution (TOPSIS) has been used for color formulation decision-making \citep{zhou2025}. CMC color difference and metamerism index were also used as optimization targets \citep{zhou2025, souissi2024color, souissi2024dyeing}. For instance, researchers have found the efficacy of recurrent neural networks (RNNs), long short-term memory (LSTM), and tree-based algorithms \citep{zhang2021, xie2024}.
	
	Modular artificial neural networks, neuro-fuzzy methods, genetic algorithms, and metaheuristic techniques such as ant colony optimization (ACO) were implemented to optimize the neural network training under non-linear conditions \citep{shamsnateri2018, yang2018, chaouch2020, Chaouch2019ant, zhu2025}. \citet{yang2018} showed that modular artificial neural networks (MANNs) improve average color difference error. 
	
	Previous studies often relied on fragmented modelling. Most of the studies used either color prediction or recipe prediction \citep{xie2024,zhang2021,yu2021,qin2021}. But we have incorporated forward quality prediction, inverse recipe reconstruction, historical retrieval, and candidate screening into one workflow, which bridges a critical gap in industrial workflows. Moreover, previous researchers used a few covariates which couldn't retain the intricate dependency of color shades properly in different steps of the dyeing process \citep{yu2021, vadood2022, xie2024, zhang2021}. It has been addressed by employing necessary industrial dependency and context in this study, such as recipe composition, process conditions, fabric information, reflectance, K/S, and engineered spectral features. Furthermore, the previous studies only explored limited multi-dye inverse prediction. Most experiments focused on only one dye or a fixed small dye combination \citep{yu2021,zhang2021,li2022optimal}. Our study with 36 dye components adds a novel contribution. Prior works report only deterministic point estimates without uncertainty bounds \citep{xie2024, zhu2025, zhou2025, yu2021, zhang2021, Chaouch2024}, a limitation addressed in this study through non-parametric quantile estimation and prediction-interval coverage analysis.

	\section{Proposed Hybrid Forward-Inverse Framework}
	\label{sec:methodology}
	
	\subsection{Problem Formulation and Architectural Overview}
	\label{sec:problem_formulation}
	
	\begin{table}[H]
		\centering
		\caption{Notations used in this study.}
		\label{tab:notation}
		\footnotesize
		\renewcommand{\arraystretch}{1.05}
		
		\begin{tabular}{
				@{}
				p{2.1cm}
				@{\hspace{0.3cm}}
				p{\dimexpr\textwidth-2.4cm\relax}
				@{}
			}
			\toprule
			\textbf{Symbol} & \textbf{Definition} \\
			\midrule
			
			\(i,\ j,\ \lambda\) &
			Indices for observations, dye components, and spectral wavelengths, respectively. \\
			
			\(N,\ D,\ K\) &
			Numbers of observations, active dye variables, and cross-validation folds; \(D=36\) and \(K=5\). \\
			
			\(\mathbf{c}_i\) &
			Observed dye-recipe concentration vector for observation \(i\). \\
			
			\(c_{ij}\) &
			Concentration of dye \(j\) in recipe \(i\), expressed as percentage on weight of fiber. \\
			
			\(T_i\) &
			Total dye concentration, \(T_i=\sum_{j=1}^{D}c_{ij}\). \\
			
			\(R_{i\lambda}\) &
			Spectral reflectance of observation \(i\) at wavelength \(\lambda\). \\
			
			\(\left(K/S\right)_{i\lambda}\) &
			Kubelka--Munk transformation of \(R_{i\lambda}\). \\
			
			\(\Delta E_{\mathrm{D65}},\Delta E_{\mathrm{F11}}\) &
			Colour differences measured under D65 and F11 illuminants. \\
			
			\(\Delta E_{\max}\) &
			Maximum illuminant-specific color difference. \\
			
			\(M_i\) &
			Absolute metamerism gap,
			\(M_i=\left|\Delta E_{\mathrm{D65},i}-\Delta E_{\mathrm{F11},i}\right|\). \\
			
			\(\mathbf{x}_i^{F},y_i^{F}\) &
			Forward-model input vector and observed color-quality response. \\
			
			\(f(\cdot),\hat{y}_i^{F}\) &
			Forward prediction function and predicted response. \\
			
			\(\mathbf{x}_i^{I},g(\cdot)\) &
			Inverse-model input vector and inverse prediction function. \\
			
			\(\hat{\mathbf{c}}_i,\mathbf{c}_i^{*}\) &
			Raw inverse prediction and sparsity-constrained recipe. \\
			
			\(\mathcal{F}(\cdot)\) &
			Sparsity post-processing function. \\
			
			\(\hat{y}_i^{CL}\) &
			Closed-loop surrogate quality prediction. \\
			
			\(s(r_q,r_c)\) &
			Ranking score between query recipe \(r_q\) and candidate recipe \(r_c\). \\
			
			\(\mathcal{R}_k(r_q)\) &
			Set of Top-\(k\) historical candidates retrieved for query \(r_q\). \\
			
			\(S_m\) &
			Composite inverse-model ranking score for model \(m\). \\
			
			\(d_i,d_r\) &
			Paired absolute-error difference at observation and recipe-group levels. \\
			
			\(n_r\) &
			Number of observations in recipe group \(r\). \\
			
			\(q_{\alpha}\) &
			Estimated response quantile at probability level \(\alpha\). \\
			
			\(\mathrm{MAE},\mathrm{RMSE}\) &
			Mean absolute error and root mean squared error. \\
			
			\(\rho\) &
			Spearman rank-correlation coefficient. \\
			
			\bottomrule
		\end{tabular}
	\end{table}
	
	Let $\mathbf{c}_i \in \mathbb{R}_{\ge 0}^D$ represent the dye-concentration vector for observation $i$ across $D$ available dyes ($D = 36$). $\mathbf{c}_i$ can be represented as $\mathbf{c}_i = [c_{i1}, c_{i2}, \dots, c_{iD}] \label{eq:recipe_vector}$ and the total dye load is given by $T_i = \sum_{j=1}^D c_{ij} \label{eq:total dye concentration}$. But a shade can be produced using a number of formulations of various dyes \citep{chaouch2020, zhou2025, cho_2026}. However, industries prefer to a maximum of $q$ active dyes \citep{Samanta2024,qin2021,xie2024, Tam2024}; (where $q = 3 \ll D$ for our study) with a minimum concentration threshold $\tau$. In this study, $\tau = 0.01\%\,\text{OWF}$ along with sparsity conditions $\|\mathbf{c}_i\|_0 \le q$ and $c_{ij} \ge \tau \cdot \mathbb{I}(c_{ij} > 0)$. For each observation $i$, the input vector to the forward quality model $\mathbf{x}_i^F$ can be expressed as equation~\ref{forward-quality-model}. A summary of the main mathematical symbols and notations used throughout this framework is provided in Table~\ref{tab:notation}. 
	
	\begin{equation}
		\mathbf{x}_i^F = \left[ \mathbf{c}_i, \, \mathbf{p}_i, \, \mathbf{s}_i \right]
		\label{forward-quality-model}
	\end{equation}
	
	where $\mathbf{p}_i$ represents production process parameters (salt, soda ash, GSM, liquor ratio, and fabric type); $\mathbf{s}_i = [R_{i,400\text{nm}}, \dots, R_{i,750\text{nm}}, \, (K/S)_{i,400\text{nm}}, \dots, (K/S)_{i,750\text{nm}}]$ is the target shade spectral profile comprising reflectance and Kubelka--Munk values across 400--750~nm. The forward model then maps this feature vector to the four shade-quality targets expressed as equation~\ref{target} where $f$ denotes to fitted target-specific model.
	
	\begin{equation}
		\hat{\mathbf{y}}_i^F
		=
		\left[
		f_{\mathrm{D65}}(\mathbf{x}_i^F),
		f_{\mathrm{F11}}(\mathbf{x}_i^F),
		f_{\max}(\mathbf{x}_i^F),
		f_{\mathrm{MG}}(\mathbf{x}_i^F)
		\right].
		\label{target}
	\end{equation}
	
	The inverse problem is the converse of forward prediction. The target shade spectral profile is defined as  $\mathbf{s}_i$ and available process context $\mathbf{p}_i$. But \emph{without} any recipe observation, the goal is to recover the dye-concentration vector $\mathbf{c}_i$. The inverse input vector is $\mathbf{x}_i^I = \left[ \mathbf{s}_i, \, \mathbf{p}_i \right]$ therefore inverse recipe reconstruction defined as equation~\ref{inverse}
	
	\begin{equation}
		\hat{\mathbf{c}}_i = g\left(\mathbf{x}_i^I\right)
		\label{inverse}
	\end{equation}
	
	\begin{figure}[H]
		\centering
		\includegraphics[width=\textwidth]{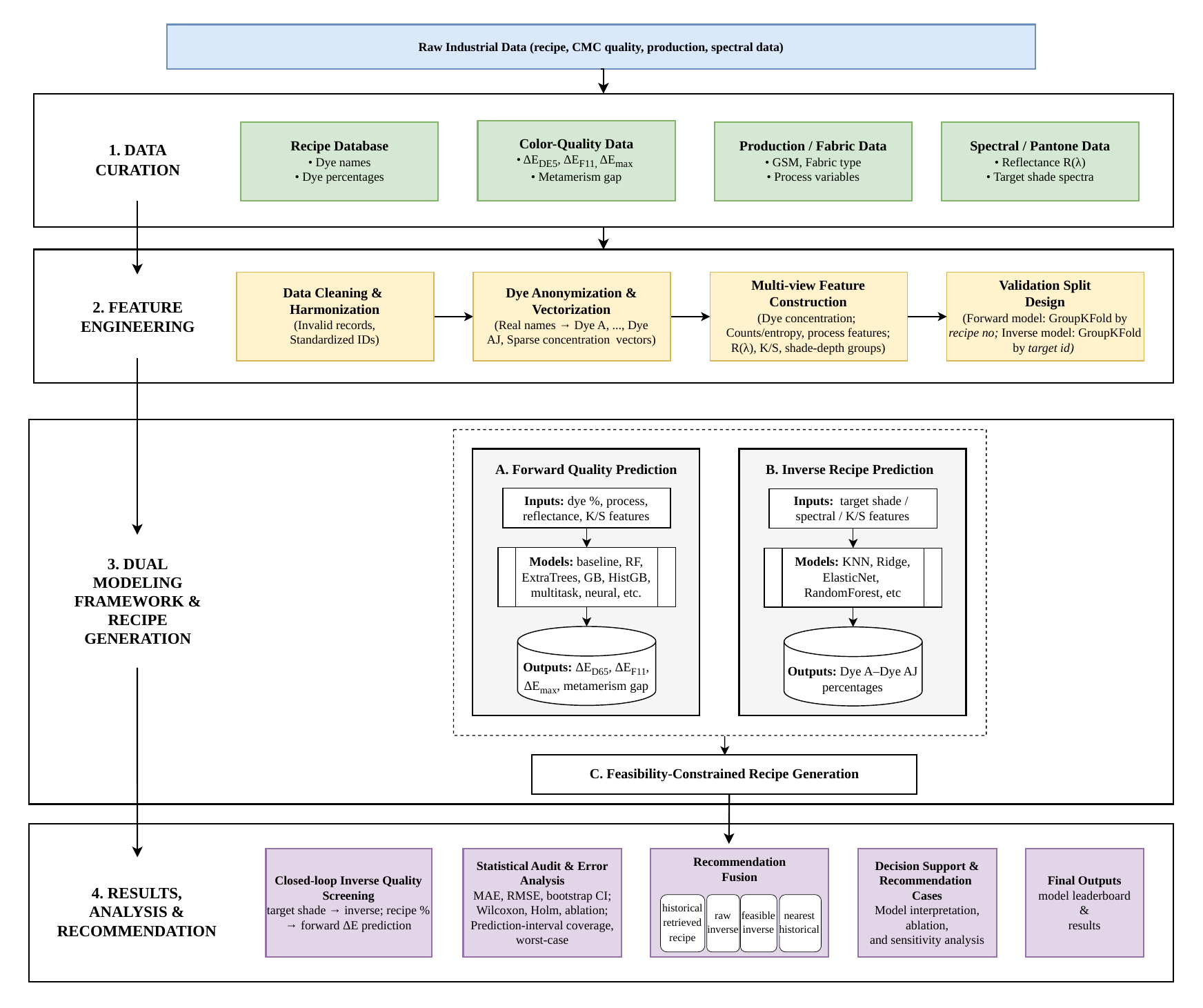}
		\caption{Overview of the data preparation, model development, validation, and decision-support workflow.}
		\label{fig:methodology_workflow}
	\end{figure}
	
	\subsection{Module A: Multi-Target Forward Quality Predictor with Uncertainty Quantification}
	\label{sec:module_a_forward}
	
	\subsubsection{Multi-Target Quality Prediction}
	The forward task was defined as supervised regression from recipe, process, dye, target reflectance, target K/S, and spectral features to four shade-quality targets can be expressed as equation~\ref{eq:forward_quality}
	\begin{equation}
		y_i^{F} =
		\left[
		\Delta E_{\mathrm{D65}},
		\Delta E_{\mathrm{F11}},
		\Delta E_{\max},
		\left|\Delta E_{\mathrm{D65}} - \Delta E_{\mathrm{F11}}\right|
		\right]
		\label{eq:forward_quality}
	\end{equation}
	where \(\Delta E_{\mathrm{D65}}\) and \(\Delta E_{\mathrm{F11}}\) are illuminant-specific color differences, \(\Delta E_{\max}\) is the worst-case color difference, and \(\left|\Delta E_{\mathrm{D65}} - \Delta E_{\mathrm{F11}}\right|\) is the metamerism gap. Separate target-specific models were fitted for the four outcomes. The absolute metamerism gap was calculated from the observed illuminant-specific color differences.
	
	\subsubsection{Prediction-Interval and Uncertainty Analysis}
	\label{sec:uncertainty_quantification}
	Nominal 80\% prediction intervals were defined by the estimated 0.10 and 0.90 quantiles. Candidate methods included a fold-wise empirical-quantile baseline and random-forest quantile models \citep{Koenker1978, Meinshausen2006}. Selection first required the recipe-cluster bootstrap confidence interval for empirical coverage to include 0.80. Eligible models were then ranked by absolute coverage error, mean pinball loss, median-prediction MAE, and mean interval width.
	
	Confidence intervals for empirical coverage, median-prediction MAE, and mean interval width were estimated using 2,000 recipe-number cluster-bootstrap resamples \citep{Field2007}. The relationship between interval width and absolute prediction error was assessed using Spearman rank correlation. For the risk-coverage analysis, observations were ranked by interval width, and tied observations were retained together.
	
	\subsection{Module B: Sparse High-Dimensional Inverse Recipe Reconstructor}
	\label{sec:module_b_inverse}
	
	\subsubsection{Inverse dye-recipe prediction}
	
	 The inverse prediction involved multi-output regression from target shade information to dye concentration percentages. The inputs are target reflectance features, target K/S features, spectral descriptors, and optional process context. The inverse prediction is defined as \(\hat{\mathbf{c}}_i\) with regression model \(g(\cdot)\) where \(\mathbf{x}_i^{I}\) is the inverse-model input for target shade \(i\) in equation~\ref{eq:inverse}. Five inverse model families were evaluated: K-nearest neighbours, Ridge regression, ElasticNet, Random Forest, and Extra Trees. Inverse models were ranked using a composite score can be expressed as equation~\ref{eq:inverse_composite_score} and formalized in Algorithm~\ref{alg:inverse_recipe}.
	
	\begin{equation}
		\hat{\mathbf{c}}_i = g\left(\mathbf{x}_i^{I}\right)
		\label{eq:inverse}
	\end{equation}
	\begin{equation}
		S_m
		=
		\mathrm{MAE}_{\mathrm{recipe},m}
		+
		0.1\,\mathrm{MAE}_{\mathrm{total},m}
		+
		\left(1-F_{1,m}^{\mathrm{active}}\right),
		\label{eq:inverse_composite_score}
	\end{equation}
	
	where \(S_m\) is the composite ranking score for model \(m\), \(\mathrm{MAE}_{\mathrm{recipe},m}\) is the recipe-vector MAE, \(\mathrm{MAE}_{\mathrm{total},m}\) is the total-dye absolute error, and \(F_{1,m}^{\mathrm{active}}\) is the active-dye F1 score. Lower values of \(S_m\) indicate better overall inverse-model performance.
	
	\begin{algorithm}[H]
		\caption{Inverse Recipe Prediction and Model Selection}
		\label{alg:inverse_recipe}
		\footnotesize
		\begin{algorithmic}[1]
			\Require Inverse input matrix
			$\mathbf{X}^{I}\in\mathbb{R}^{N\times p}$,
			observed recipe matrix
			$\mathbf{C}\in\mathbb{R}^{N\times D}$,
			grouped folds
			$\{(\mathcal{I}_{k}^{\mathrm{tr}},
			\mathcal{I}_{k}^{\mathrm{va}})\}_{k=1}^{K}$,
			model set $\mathcal{M}_{\mathrm{inv}}$,
			sparsity map $\mathcal{F}_{\tau,q}$
			
			\Ensure Selected model $m^{*}$ and sparse out-of-fold predictions
			$\hat{\mathbf{C}}^{*}$
			
			\For{$m\in\mathcal{M}_{\mathrm{inv}}$}
			\State Initialize
			$\hat{\mathbf{C}}^{(m)}
			\in\mathbb{R}^{N\times D}$
			
			\For{$k=1,\ldots,K$}
			\State Fit
			$m:
			\mathbf{X}^{I}_{\mathcal{I}_{k}^{\mathrm{tr}}}
			\mapsto
			\mathbf{C}_{\mathcal{I}_{k}^{\mathrm{tr}}}$
			
			\State
			$\hat{\mathbf{C}}_{\mathcal{I}_{k}^{\mathrm{va}}}^{(m)}
			\gets
			m\!\left(
			\mathbf{X}^{I}_{\mathcal{I}_{k}^{\mathrm{va}}}
			\right)$
			\EndFor
			
			\For{$i=1,\ldots,N$}
			\State
			$\hat{\mathbf{c}}_{i}^{*(m)}
			\gets
			\mathcal{F}_{\tau,q}
			\!\left(
			\hat{\mathbf{c}}_{i}^{(m)}
			\right)$
			\EndFor
			
			\State
			$\mathrm{MAE}_{\mathrm{recipe}}^{(m)}
			\gets
			\frac{1}{ND}
			\sum_{i=1}^{N}
			\sum_{j=1}^{D}
			\left|
			c_{ij}-\hat{c}_{ij}^{*(m)}
			\right|$
			
			\State
			$\mathrm{MAE}_{\mathrm{total}}^{(m)}
			\gets
			\frac{1}{N}
			\sum_{i=1}^{N}
			\left|
			\sum_{j=1}^{D}c_{ij}
			-
			\sum_{j=1}^{D}\hat{c}_{ij}^{*(m)}
			\right|$
			
			\State Compute
			$F_{1,\mathrm{active}}^{(m)}$
			
			\State
			$S_m
			\gets
			\mathrm{MAE}_{\mathrm{recipe}}^{(m)}
			+
			0.1\,\mathrm{MAE}_{\mathrm{total}}^{(m)}
			+
			\left(
			1-F_{1,\mathrm{active}}^{(m)}
			\right)$
			\EndFor
			
			\State
			$m^{*}\gets
			\arg\min_{m\in\mathcal{M}_{\mathrm{inv}}}S_m$
			
			\State
			$\hat{\mathbf{C}}^{*}
			\gets
			\hat{\mathbf{C}}^{*(m^{*})}$
		\end{algorithmic}
	\end{algorithm}
	
	\subsubsection{Inverse Recipe Meta-model}
	
	An inverse meta-model was designed and evaluated in the second stage using the out-of-fold predictions of the ten base inverse-model configurations, described in algorithm~\ref{alg:hybrid_ensemble}. Base predictions were converted into component-level meta-features, including the predictions from each base model, their mean and median, recipe-level predicted totals, maximum concentrations, and active-component counts.
	
	\begin{algorithm}[H]
		\caption{Fold-Preserving Meta-Model Blending for Inverse Recipe Reconstruction}
		\label{alg:hybrid_ensemble}
		\footnotesize
		\begin{algorithmic}[1]
			\Require Sparse out-of-fold base predictions
			$\{\hat{\mathbf{C}}^{(b)}\}_{b=1}^{B}$,
			observed recipes $\mathbf{C}\in\mathbb{R}^{N\times D}$,
			grouped folds
			$\{(\mathcal{I}_{k}^{\mathrm{tr}},\mathcal{I}_{k}^{\mathrm{va}})\}_{k=1}^{K}$,
			ensemble set
			$\mathcal{H}=\{\mathrm{mean},\mathrm{median},\mathrm{Ridge},\mathrm{ElasticNet}\}$,
			sparsity map $\mathcal{F}_{\tau,q}$ with $\tau=0.01$ and $q=3$
			
			\Ensure Selected predictions $\hat{\mathbf{C}}^{*}$
			
			\For{$i=1,\ldots,N$ and $j=1,\ldots,D$}
			\State
			$\mathbf{z}_{ij}\gets
			\left[
			\{\hat{c}_{ij}^{(b)}\}_{b=1}^{B},
			\operatorname{mean}_{b}\hat{c}_{ij}^{(b)},
			\operatorname{median}_{b}\hat{c}_{ij}^{(b)},
			\{\hat{T}_{i}^{(b)},\hat{M}_{i}^{(b)},\hat{A}_{i}^{(b)}\}_{b=1}^{B}
			\right]$
			\State
			$\hat{T}_{i}^{(b)}=\sum_{j=1}^{D}\hat{c}_{ij}^{(b)}$,
			$\hat{M}_{i}^{(b)}=\max_{j}\hat{c}_{ij}^{(b)}$,
			$\hat{A}_{i}^{(b)}=\sum_{j=1}^{D}
			\mathbb{I}\!\left(\hat{c}_{ij}^{(b)}\geq\tau\right)$
			\EndFor
			
			\For{$h\in\mathcal{H}$}
			\If{$h\in\{\mathrm{mean},\mathrm{median}\}$}
			\State
			$\hat{c}_{ij}^{(h)}
			\gets
			h\!\left(
			\hat{c}_{ij}^{(1)},\ldots,\hat{c}_{ij}^{(B)}
			\right)$
			\Else
			\For{$k=1,\ldots,K$}
			\State Fit $h$ on
			$\{(\mathbf{z}_{ij},c_{ij}):
			i\in\mathcal{I}_{k}^{\mathrm{tr}},
			j=1,\ldots,D\}$
			\State Predict $\hat{c}_{ij}^{(h)}$ for
			$i\in\mathcal{I}_{k}^{\mathrm{va}}$
			\EndFor
			\EndIf
			
			\For{$i=1,\ldots,N$}
			\State
			$\hat{\mathbf{c}}_{i}^{*(h)}
			\gets
			\mathcal{F}_{\tau,q}
			\!\left(\hat{\mathbf{c}}_{i}^{(h)}\right)$
			\EndFor
			
			\State
			$S_h\gets
			\mathrm{MAE}_{\mathrm{recipe}}^{(h)}
			+0.1\,\mathrm{MAE}_{\mathrm{total}}^{(h)}
			+\left(1-F_{1,\mathrm{active}}^{(h)}\right)$
			\EndFor
			
			\State
			$h^{*}\gets\arg\min_{h\in\mathcal{H}}S_h$
			\State
			$\hat{\mathbf{C}}^{*}\gets\hat{\mathbf{C}}^{*(h^{*})}$
		\end{algorithmic}
	\end{algorithm}
	
	Ridge, ElasticNet, arithmetic-mean blending, and component-wise median blending were evaluated. Meta-model validation preserved the original held-out fold assignments: for each fold, the trainable meta-models were fitted using base-model out-of-fold predictions from the remaining folds and evaluated on the withheld fold. The mean and median blends required no fitting. Sparse post-processing was then applied using the same threshold and maximum-component rules as the base inverse models.
	
	\subsubsection{Sparsity-Constrained Inverse Recipe Post-processing}
	
	Raw inverse predictions were converted into sparse candidate vectors by clipping negative values to zero (presented in equation~\ref{eq:inverse_sparse_postprocessing}), removing components below 0.01\% OWF to zero. Then it retained the three largest predicted components (top (k), where k = 3). The retained concentrations were rescaled to preserve the predicted total dye load.
	\begin{equation}
		\hat{c}_{ij}^{+}
		=
		\max\left(\hat{c}_{ij},0\right),
		\qquad
		\mathbf{c}_i^{*}
		=
		\mathcal{F}\left(\hat{\mathbf{c}}_i^{+}\right),
		\label{eq:inverse_sparse_postprocessing}
	\end{equation}
	where \(\mathcal{F}(\cdot)\) is the sparsity post-processing function and \(\mathbf{c}_i^{*}\) is the sparsified inverse-recipe proposal.
	
	\subsection{Module C: Feasibility-Constrained Selection and Closed-Loop Auditing Layer}
	\label{sec:module_c_selection}
	
	\subsubsection{Retrospective Historical Recipe Retrieval}
	
	Historical recipe retrieval was evaluated under three exclusion rules. They are (i) self-exclusion: only the query observation was removed; (ii) recipe-group-safe retrieval: all observations sharing the query recipe number were excluded; and (iii) strict unseen-target retrieval: observations sharing the query target identifier were excluded.
	
	Three retrieval profiles were evaluated: a recipe-process nearest-neighbour reference, a spectral-process profile incorporating the candidate’s observed \(\Delta E_{\max}\), and a balanced recipe-spectral-quality profile including recipe descriptors and a total-dye-load penalty. In Algorithm~\ref{alg:recommendation}, \(A_p(r_i,r_c)\) indicates whether candidate \(r_c\) is eligible for query \(r_i\) under exclusion protocol \(p\).
	
	Missing feature values were replaced by their median, and the retrieval features were standardized before calculating Euclidean distance. For each
	query, the candidate distances, observed candidate \(\Delta E_{\max}\), and total dye loads were separately min-max scaled within the eligible candidate set. The retrieval score was defined as equation~\ref{eq:retrieval_score}.
	
	\begin{equation}
		s_{ic}^{(h)}
		=
		w_{d}^{(h)}\widetilde{d}_{ic}
		+
		w_{e}^{(h)}\widetilde{\Delta E}_{\max,c}
		+
		w_{T}^{(h)}\widetilde{T}_{c},
		\label{eq:retrieval_score}
	\end{equation}
	
	where \(i\), \(c\), and \(h\) represent the query, historical candidate, and retrieval profile, respectively. \(\widetilde{d}_{ic}\), \(\widetilde{\Delta E}_{\max,c}\), and \(\widetilde{T}_c\) denote the scaled Euclidean distance, observed candidate \(\Delta E_{\max}\), and total dye load. Their respective profile weights are \(w_d^{(h)}\), \(w_e^{(h)}\), and \(w_T^{(h)}\). The weights \((w_d,w_e,w_T)\) were \((1.00,0,0)\) for the recipe-process, \((0.75,0.25,0)\) for the spectral-process profile, and \((0.70,0.25,0.05)\) for the balanced recipe-spectral-quality profile.
	
	\begin{algorithm}[H]
		\caption{Retrospective Outcome-Informed Top-\(k\) Recipe Retrieval}
		\label{alg:recommendation}
		\footnotesize
		\begin{algorithmic}[1]
			\Require Query set \(\mathcal{Q}\), candidate library \(\mathcal{C}\), retrieval profile \(h\), exclusion protocol \(p\), observed candidate quality \(e_c=\Delta E_{\max,c}\), retrieval depths \(\mathcal{K}=\{1,3,5\}\)
			\Ensure Query-level scores \(e_{i,k}^{(p,h)}\) and mean retrieval performance \(\bar{e}_{k}^{(p,h)}\)
			
			\For{each query \(r_i\in\mathcal{Q}\)}
			\State
			\(\mathcal{C}_{i}^{(p)}
			\gets
			\{r_c\in\mathcal{C}:
			A_p(r_i,r_c)=1\}\)
			
			\State
			\(\mathbf{z}_i\gets\phi_h(r_i)\)
			
			\For{each \(r_c\in\mathcal{C}_{i}^{(p)}\)}
			\State
			\(\mathbf{z}_c\gets\phi_h(r_c)\)
			
			\State
			\(s_{ic}^{(h)}
			\gets
			s_h(\mathbf{z}_i,\mathbf{z}_c,e_c)\)
			\EndFor
			
			\State
			\(\pi_i
			\gets
			\operatorname{argsort}_{r_c\in\mathcal{C}_{i}^{(p)}}
			s_{ic}^{(h)}\)
			
			\For{each \(k\in\mathcal{K}\)}
			\State
			\(\mathcal{R}_{i,k}^{(p,h)}
			\gets
			\{\pi_i(1),\ldots,\pi_i(k)\}\)
			
			\State
			\(e_{i,k}^{(p,h)}
			\gets
			\min_{r_c\in\mathcal{R}_{i,k}^{(p,h)}}
			\Delta E_{\max,c}\)
			\EndFor
			\EndFor
			
			\For{each \(k\in\mathcal{K}\)}
			\State
			\(\displaystyle
			\bar{e}_{k}^{(p,h)}
			\gets
			\frac{1}{|\mathcal{Q}|}
			\sum_{i=1}^{|\mathcal{Q}|}
			e_{i,k}^{(p,h)}\)
			\EndFor
		\end{algorithmic}
	\end{algorithm}

	\subsubsection{Closed-Loop Surrogate Quality Screening}
	
	The sparsified inverse-recipes were evaluated by passing them through the fitted forward models. Recipe-derived variables were recomputed from each proposed dye vector. The query’s target spectral and available process descriptors were retained. The resulting feature vector was passed through the final-refit forward models to obtain surrogate predictions of $\Delta E_{\mathrm{D65}}$, $\Delta E_{\mathrm{F11}}$, $\Delta E_{\max}$, and the absolute metamerism gap. The entire process can be written as equation~\ref{eq:closed_loop_workflow}
	
	\begin{equation}
		\begin{aligned}
			\hat{\mathbf{c}}_i
			&=
			g\left(\mathbf{x}_i^{I}\right),\\
			\mathbf{c}_i^{*}
			&=
			\mathcal{F}\left(\hat{\mathbf{c}}_i\right),\\
			\hat{y}_i^{CL}
			&=
			f\left(\mathbf{x}_i^{F}\left(\mathbf{c}_i^{*}\right)\right).
		\end{aligned}
		\label{eq:closed_loop_workflow}
	\end{equation}
	
	where \(g(\cdot)\) is the inverse recipe model, \(\mathcal{F}(\cdot)\) is the feasibility constraint function, and \(f(\cdot)\) is the forward quality model. The resulting \(\hat{y}_i^{CL}\) represents the predicted shade-quality outcome. Candidate recipes were screened using predefined surrogate acceptance thresholds. The thresholds were \(\Delta E_{\mathrm{D65}} \leq 1.0\), \(\Delta E_{\mathrm{F11}} \leq 1.0\), \(\Delta E_{\max} \leq 1.0\), and absolute metamerism gap \(\leq 0.5\). 
	
	\subsubsection{Historical and inverse recommendation fusion}
	
	Inverse-generated recipes were compared with historical recipe retrieval. Four candidate sources were used: historical recipes, raw inverse predictions, sparsity-constrained inverse predictions, and the nearest historical recipe to the sparse inverse prediction. Historical retrieval search for similar previous shades. Inverse prediction generates a dye vector from the target shade. The nearest-historical-to-inverse candidate grounds the model output in an existing industrial recipe. This fusion provides both new data-driven candidates and historically validated alternatives presented as equation~\ref{historical}, where \(\mathcal{C}_i\) is the candidate set for query shade \(i\).
	
	\begin{equation}
		\label{historical}
		\mathcal{C}_i =
		\left\{
		\mathbf{c}_{\mathrm{hist}},
		\hat{\mathbf{c}}_{\mathrm{raw}},
		\mathbf{c}^{*}_{\mathrm{feasible}},
		\mathbf{c}_{\mathrm{nearest\text{-}hist}}
		\right\}
	\end{equation}
	
	\subsection{Case Study Data and Leakage-Controlled Validation Protocol}
	\label{sec:case_study_validation}
	\subsubsection{Materials and Data Description}
	\label{sec:materials}
	
	The dataset used in this study was collected from an industrial textile-dyeing partner in Bangladesh under a non-disclosure agreement for sharing sensitive commercial and industry identifier information. The data represent laboratory-to-bulk shade development records in which dye recipes were prepared to match buyer-specified PANTONE target shades and were evaluated using color-difference quality metrics. The dataset consists of four parts: (a) production and process-related parameters, (b) target shade spectral records, (c) dye recipes, and (d) post-approval CMC results. Initially, the recipe table included 259 concentration records, 101 recipes, and 63 target shades. Each recipe included up to three dye components. Dye concentration was recorded as percentage on weight of fiber (\% OWF). The color-quality part contained illuminant-level observations measured under D65 daylight and F11 fluorescent illuminants. The production part included process variables such as fabric type, knitting structure, GSM, fiber composition, liquor ratio, salt, soda ash, and trial count. The spectral part contained PANTONE target reflectance values from 400--750 nm at 10 nm intervals.
	
	After cleaning, harmonization, target matching, and removal of incomplete cases, the final dataset contained 212 supervised observations, 85 recipe numbers, and 51 target shades. Four target variables were used: \(\Delta E_{\mathrm{D65}}\), \(\Delta E_{\mathrm{F11}}\), \(\Delta E_{\max}\), and absolute metamerism gap. They illustrates daylight color difference, fluorescent-light color difference, worst-illuminant color difference, and illuminant-dependent shade instability, respectively. Due to confidentiality, commercial dye names were anonymized as Dye A to Dye AJ and it produced 36 active dye variables. Each recipe was encoded as a sparse dye-concentration vector as described in equation~\ref{eq:recipe_vector}. Target shades were represented using reflectance features and Kubelka-Munk K/S features. The final feature matrix included recipe-composition, anonymized dye-vector, dye-family, process, categorical fabric, reflectance, K/S, and spectral descriptor features for forward quality prediction and inverse dye-recipe modelling.
	
	\subsubsection{Data Preprocessing}
	
	Recipe records were standardized by normalizing identifiers and dye names. The CMC records were harmonized and retained total color difference with lightness-, chroma-, and hue-related components \citep{Luo1986, Oulton2017}. Shade approval was evaluated under D65 and F11. The recipe, quality, production, reflectance, and K/S features were then derived, recipe, and target identifiers. Identifier variables were excluded from model input features to avoid leakage. Missing numerical values within the modelling pipeline were handled using training-fold-based imputation. Each dye recipe was converted into a sparse numerical concentration vector defined in Sec~\ref{sec:problem_formulation}. Recipe-level dye count, total dye concentration, maximum dye component, dye-component entropy, and dye-ratio descriptors were also computed. The total dye concentration defined in the same Sec~\ref{sec:problem_formulation}. Dye-component entropy quantified as equation~\ref{eq:dye_entropy}. It measures how evenly dye concentration was partitioned across active recipe components.
	\begin{equation}
		H_i = -\sum_{j: p_{ij} > 0} p_{ij} \ln(p_{ij}), \qquad p_{ij} = \frac{c_{ij}}{\sum_{m=1}^D c_{im}},
		\label{eq:dye_entropy}
	\end{equation}
	
	\subsubsection{Exploratory Data Analysis and Feature Engineering}
	\begin{figure}[H]
		\centering
		\includegraphics[width=\textwidth]{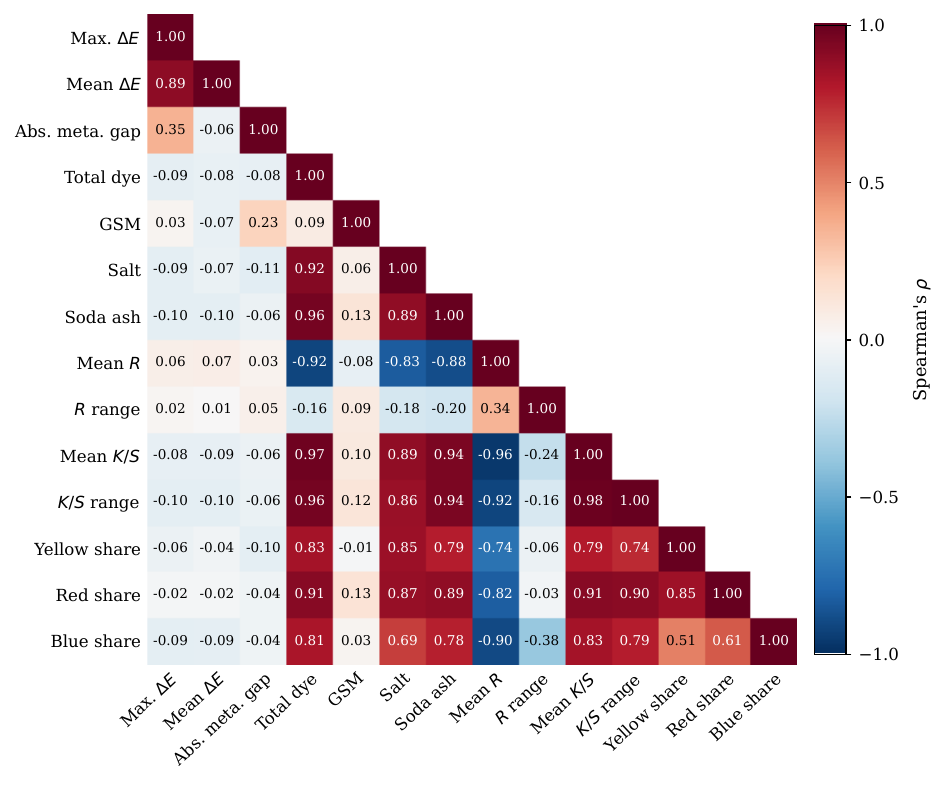}
		\caption{Spearman correlation matrix for the selected features and target variables.}
		\label{fig:core_spearman}
	\end{figure}
	The entire candidate set contained 222 predictors: 36 dye-concentration variables, 16 dye-composition descriptors, 12 recipe-related variables, 26 process variables, 30 one-hot-encoded categorical variables, 36 reflectance values, 36 Kubelka--Munk (K/S) values, and 30 spectral descriptors. For reflectance \(R_{i\lambda}\), expressed as a proportion between zero and one, the Kubelka--Munk transformation was calculated as equation~\eqref{eq:km}.
	
	\begin{equation}
		\label{eq:km}
		\left(\frac{K}{S}\right)_{i\lambda}
		=
		\frac{\left(1-R_{i\lambda}\right)^2}
		{2R_{i\lambda}}.
	\end{equation}
	
	The spectral variables covered 400--750~nm at 10~nm intervals. The descriptors included the mean, standard deviation, minimum, maximum, range, area, slope, derivative statistics, and wavelength-band means. Spearman correlation screening (presented in figure~\ref{fig:core_spearman}) and variance inflation factor (VIF) diagnostics were used to assess multicollinearity.
	
	\subsubsection{Feature Selection}
	
\begin{figure}[H]
		\centering
		\includegraphics[width=\textwidth]{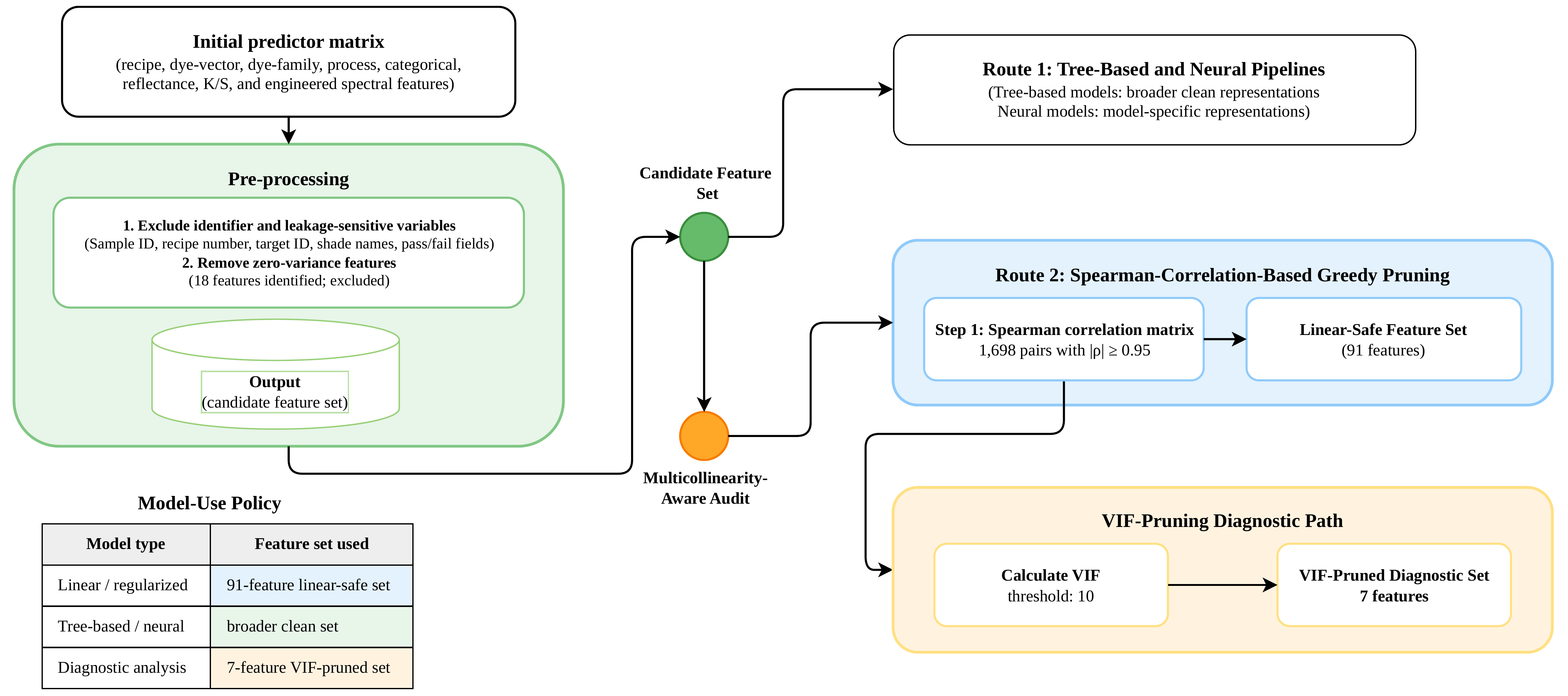}
		\caption{Feature preprocessing and multicollinearity-aware feature-selection workflow.}
		\label{fig:feature_selection_workflow}
	\end{figure}

	Highly correlated predictors were identified using Spearman correlation (feature selection workflow presented in Fig~\ref{fig:feature_selection_workflow}). The analysis identified 1,698 predictor pairs with \(\lvert\rho\rvert \geq 0.95\). For each correlated pair, a greedy pruning retained the predictor with modelling priority. This produced a 91-feature set at \(\lvert\rho\rvert \geq 0.95\) for linear modelling. Then by applying a variance inflation factor (VIF) threshold of 10, reduced it to a 7-feature set, later used by ordinary multiple linear regression model. Ridge, Lasso, ElasticNet, and PLS used the 91-feature set. Neural models used model-specific tabular, spectral, or fused input representations.
	
	The primary forward and inverse models were evaluated using five-fold GroupKFold cross-validation grouped by recipe number. It ensures that observations from the same recipe remained within one fold. Strict inverse generalization was evaluated using five-fold GroupKFold cross-validation grouped by target identifier so that all observations associated with the same target shade remained within one fold. The 51 target identifiers were distributed across the five validation folds.
	
	\subsubsection{Hyperparameter Optimization and Model Selection}
	\label{sec:hyperparameter_optimization}
	
	Hyperparameters were optimized for each prediction target, model family, and feature representation. The tuned forward-model families were Random Forest, Extra Trees, Gradient Boosting, Histogram Gradient Boosting, Ridge, and ElasticNet, evaluated using recipe-process, spectral-only, and tree-safe-full feature sets. RandomizedSearchCV tested up to 200 configurations per model--target--feature combination using negative MAE, the predefined five recipe-grouped folds, and a fixed random seed of 42 \citep{Pedregosa2011}.
	
	Numerical imputation was fitted within each training partition, with standardization applied to Ridge and ElasticNet. Predictions were constrained to be non-negative. Selected parameters were ranked using pooled out-of-fold MAE. The selected estimators were then refitted on all supervised observations. Inverse-model hyperparameters were tuned for five model families and two input representations using the same recipe-grouped folds. Up to 200 configurations were evaluated per model--representation combination; for KNN, all 28 unique configurations were tested.
	
	\subsubsection{Evaluation metrics}
	
	Forward quality prediction was evaluated using out-of-fold regression metrics. The mean absolute error and root mean squared error penalized errors can be expressed as equations~\ref{eq:mae} and \ref{eq:rmse}. Median absolute error, bias, and Spearman rank correlation were reported as secondary forward-prediction metrics. Inverse recipe prediction was evaluated using recipe-vector MAE, total-dye absolute error, active-dye precision, active-dye recall, active-dye F1 score, and active Top-\(k\) recall.
	
	\begin{equation}
		MAE = \frac{1}{n}\sum_{i=1}^{n}\left|y_i - \hat{y}_i\right|
		\label{eq:mae}
	\end{equation}
	
	\begin{equation}
		RMSE = \sqrt{\frac{1}{n}\sum_{i=1}^{n}\left(y_i - \hat{y}_i\right)^2}
		\label{eq:rmse}
	\end{equation}

	\subsubsection{Controlled Input-Block Ablation}
	
	A controlled ablation experiment was conducted for Random Forest (RF), Gradient Boosting (GB) and Histogram Gradient Boosting (HGB) for predicting all four target variables. Recipe, process, reflectance, K/S, and engineered spectral feature blocks were removed one at a time. Recipe-process-only and spectral-only configurations were also examined. The ablation effect was defined as equation~\ref{eq:ablation}.
	
	\begin{equation}
		\Delta\mathrm{MAE}
		=
		\mathrm{MAE}_{\mathrm{ablated}}
		-
		\mathrm{MAE}_{\mathrm{full}}.
		\label{eq:ablation}
	\end{equation}
	
	Positive values indicate that removing an input block increased prediction error for the ablated model and vice versa. Confidence intervals were estimated using recipe-number cluster-bootstrap resampling. Paired comparisons were conducted at the recipe-group level, and Holm correction was applied across model–target family and globally across removals.
	
	\subsubsection{Permutation Importance and Partial Dependence}
	
	Permutation importance was calculated for RF, GB, and HGB across four targets using the tuned tree-safe-full configurations and five
	recipe-grouped folds \citep{Goldstein2015, Hastie2009}. Importance was the change in held-out MAE after permutation. Feature blocks were permuted with a common row permutation over 20 repetitions. Individual features used two repetitions. The 95\% confidence intervals used 10,000 recipe-number cluster-bootstrap resamples.
	
	For each model-target pair, the highest-importance eligible continuous feature was selected as the first PDP variable. The second was selected
	from a different feature block with \(\lvert\rho\rvert<0.90\), when available. Models were refitted on all 212 observations, and two-way PDPs
	were calculated over 18 quantile-spaced points from the 5th to the 95th percentile. 
	
	\subsubsection{Statistical Testing}
	\label{sec:statistical_testing}
	
	Statistical comparisons were based on paired out-of-fold predictions, with MAE as the primary outcome and RMSE and Spearman correlation as secondary metrics. For each observation, the candidate--reference absolute-error difference was defined as equation~\ref{eq:paired_row_error_difference}, where \(d_i<0\) indicates lower error for the candidate model. But observations are sharing a recipe number which is not independent. So, these differences were averaged within each recipe group expressed as equation~\ref{eq:recipe_group_error_difference}, where \(n_r\) is the number of observations in recipe group \(r\).
	\begin{equation}
		d_i
		=
		\left|y_i-\hat{y}_{i,\mathrm{candidate}}\right|
		-
		\left|y_i-\hat{y}_{i,\mathrm{reference}}\right|,
		\label{eq:paired_row_error_difference}
	\end{equation}
	 
	\begin{equation}
		d_r
		=
		\frac{1}{n_r}
		\sum_{i\in r} d_i,
		\label{eq:recipe_group_error_difference}
	\end{equation}
	 Forward-model comparisons therefore used 85 recipe-level differences rather than 212 observation rows. One-sided Wilcoxon signed-rank tests evaluated the alternative \(\operatorname{median}(d_r)<0\), and confidence intervals for mean error differences were estimated using 10,000 recipe-number cluster-bootstrap resamples.
	 
	 \subsubsection{Exploratory Observational Effect and Heterogeneity Analysis}
	 
	 An exploratory analysis examined associations between eight process variables and four color-quality outcomes. Each continuous exposure was dichotomized at its sample median, with observations above the median treated as the exposed group. Thirty-five numerical adjustment covariates were used after excluding identifiers, outcomes, and the evaluated exposures.
	 
	 For each exposure-outcome pair, the average effect was estimated using augmented inverse probability weighting (AIPW) \citep{Bang2005}. Propensity scores were estimated using logistic regression and restricted to the interval \([0.05,0.95]\). Separate ridge-regression outcome models were fitted for the exposed and unexposed groups. Let \(T_i\) denote exposure status, \(Y_i\) the outcome, \(e(\mathbf{X}_i)\) the estimated propensity score, and \(\hat{\mu}_t(\mathbf{X}_i)\) the estimated outcome under exposure level \(t\). The AIPW estimator was
	 
	 \begin{equation}
	 	\hat{\tau}_{\mathrm{AIPW}}
	 	=
	 	\frac{1}{n}\sum_{i=1}^{n}
	 	\left[
	 	\hat{\mu}_1(\mathbf{X}_i)-\hat{\mu}_0(\mathbf{X}_i)
	 	+
	 	\frac{T_i\{Y_i-\hat{\mu}_1(\mathbf{X}_i)\}}{e(\mathbf{X}_i)}
	 	-
	 	\frac{(1-T_i)\{Y_i-\hat{\mu}_0(\mathbf{X}_i)\}}
	 	{1-e(\mathbf{X}_i)}
	 	\right].
	 \end{equation}
	 
	 Covariate balance was assessed using standardized mean differences before and after weighting \citep{Austin2015}. Confidence intervals were obtained from 10,000 bootstrap resamples of the unit-level AIPW scores, and \(p\)-values were adjusted using the Holm procedure \citep{Holm1979}.
	 
	 Heterogeneity was explored using random-forest T-learners fitted separately to the exposed and unexposed observations \citep{Kunzel2019}. The observation-level conditional effect was defined as
	 
	 \begin{equation}
	 	\hat{\tau}_i
	 	=
	 	\hat{\mu}_1(\mathbf{X}_i)-\hat{\mu}_0(\mathbf{X}_i).
	 \end{equation}
	
	\begin{landscape}
	\section{Results and Analysis}
	\label{sec:results}

	\subsection{Module A: Forward Color Quality Prediction and Uncertainty Quantification}
	\label{sec:results_forward_uncertainty}
	
	\subsubsection{Multi-Target Forward Prediction Benchmarks}
	Performance metrics were calculated from predictions pooled across the five held-out folds. Table~\ref{tab:forward_prediction_performance} reports the model with the lowest pooled out-of-fold MAE for each target.
	
		\begin{table}[H]
			\centering
			\caption{Lowest pooled out-of-fold MAE among the directly evaluated forward and uncertainty models for each color-quality target.}
			\label{tab:forward_prediction_performance}
			\small
			\resizebox{\linewidth}{!}{%
				\begin{tabular}{l l c c c c c c}
					\toprule
					\textbf{Target}
					&
					\textbf{Model}
					&
					\textbf{MAE}
					&
					\textbf{95\% cluster-bootstrap CI}
					&
					\textbf{RMSE}
					&
					\textbf{Median AE}
					&
					\textbf{Bias}
					&
					\textbf{Spearman \(\boldsymbol{\rho}\)}
					\\
					\midrule
					
					\(\Delta E_{\mathrm{D65}}\)
					&
					HGB- (Sp.)
					&
					0.1655
					&
					0.1475--0.1853
					&
					0.2094
					&
					0.1558
					&
					0.0069
					&
					0.2368
					\\
					
					\(\Delta E_{\mathrm{F11}}\)
					&
					GB- (RP)
					&
					0.1523
					&
					0.1316--0.1759
					&
					0.1977
					&
					0.1279
					&
					0.0040
					&
					0.3988
					\\
					
					\(\Delta E_{\max}\)
					&
					HGB- (Sp.)
					&
					0.1398
					&
					0.1230--0.1596
					&
					0.1822
					&
					0.1228
					&
					0.0087
					&
					0.2507
					\\
					
					Absolute metamerism gap
					&
					QRF-LINEAR-KM (LSP)
					&
					0.1046
					&
					0.0909--0.1191
					&
					0.1364
					&
					0.0865
					&
					-0.0083
					&
					0.4746
					\\
					
					\bottomrule
				\end{tabular}%
			}
			
			\vspace{0.3cm}
			
			\begin{minipage}{0.98\linewidth}
				\footnotesize
				\textit{Note.} Confidence intervals were estimated using 10,000 recipe-number cluster-bootstrap resamples. HGB: histogram gradient
				boosting; GB: gradient boosting; QRF: quantile random forest; Sp.: spectral-only; RP: recipe-process; LSP: linear-safe plus K/S.
			\end{minipage}
		\end{table}
	\end{landscape}

	\begin{figure}[H]
		\centering
		\includegraphics[width=\textwidth]{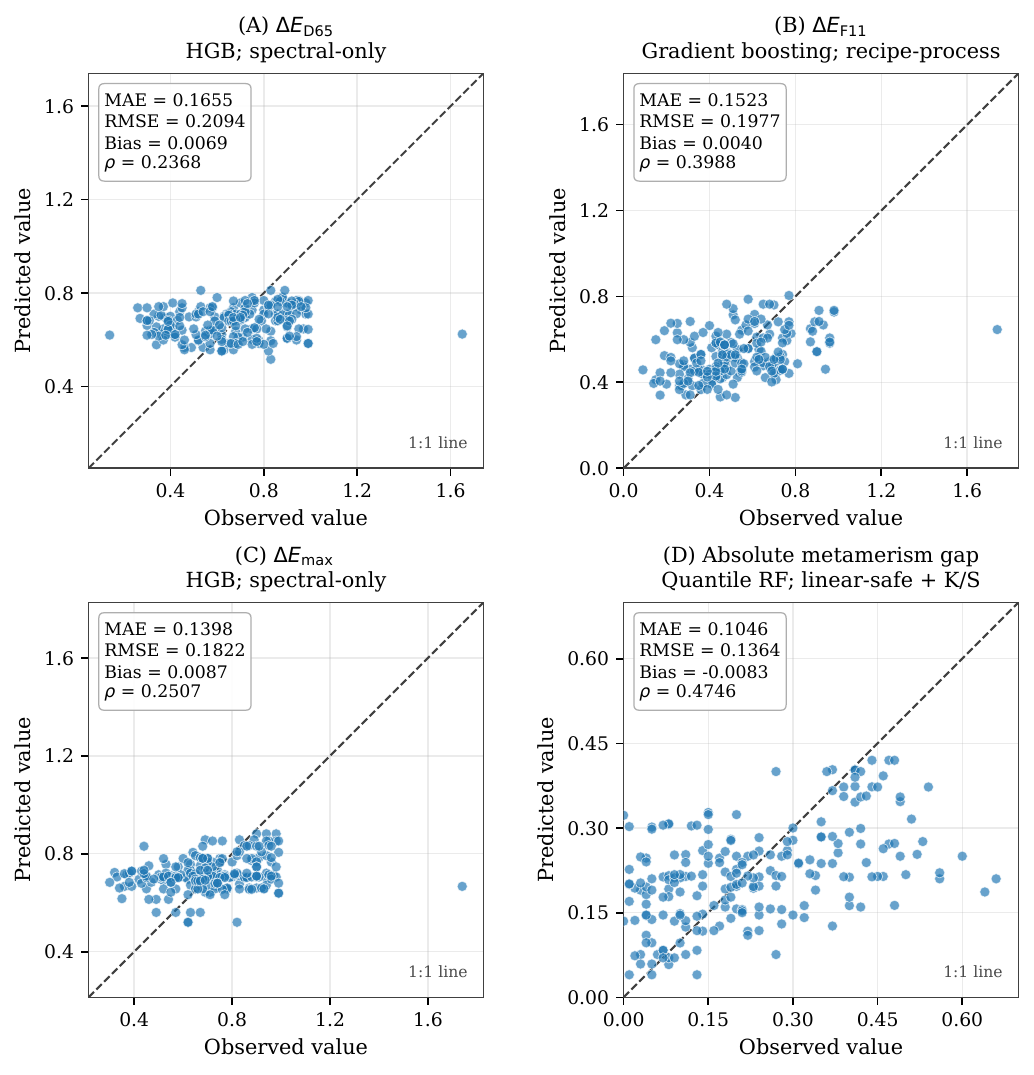}
		\caption{Observed and out-of-fold predicted values for the lowest-MAE model of each color-quality target. Panel A shows \(\Delta E_{\mathrm{D65}}\), Panel B shows \(\Delta E_{\mathrm{F11}}\), Panel C shows \(\Delta E_{\max}\), and Panel D shows the absolute metamerism gap.}
		\label{fig:forward_observed_vs_predicted}
	\end{figure}
	.
	\begin{figure}[H]
		\centering
		\includegraphics[width=0.90\textwidth]{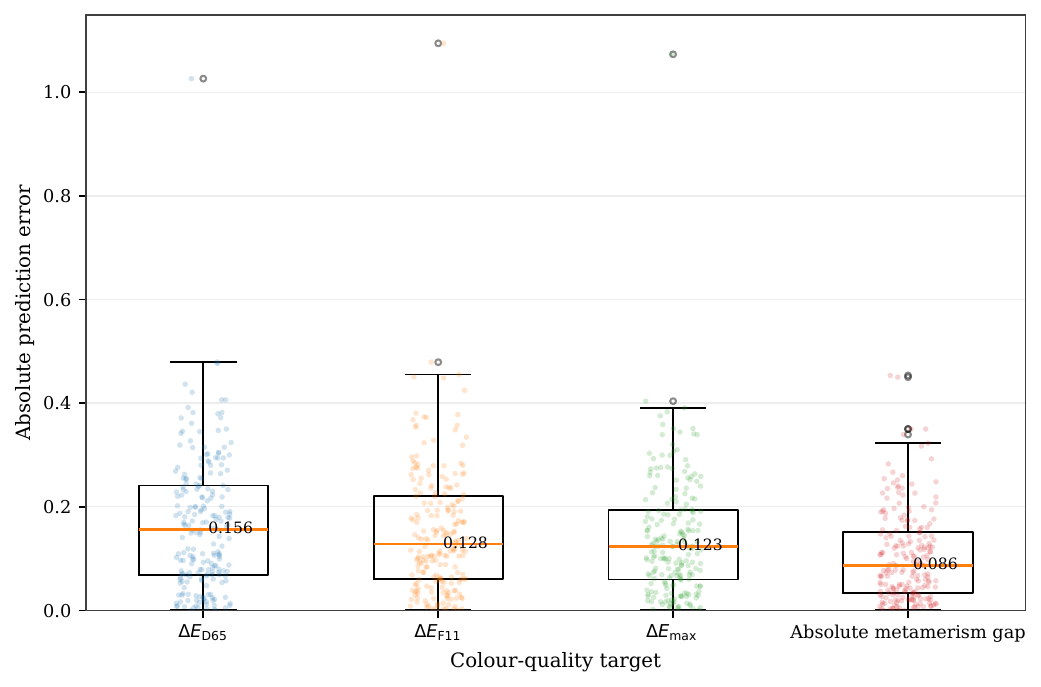}
		\caption{Distribution of absolute out-of-fold prediction errors for the lowest-MAE model of each color-quality target.}
		\label{fig:forward_absolute_error_distributions}
	\end{figure}
	
	\begin{figure}[H]
		\centering
		\includegraphics[width=0.95\textwidth]{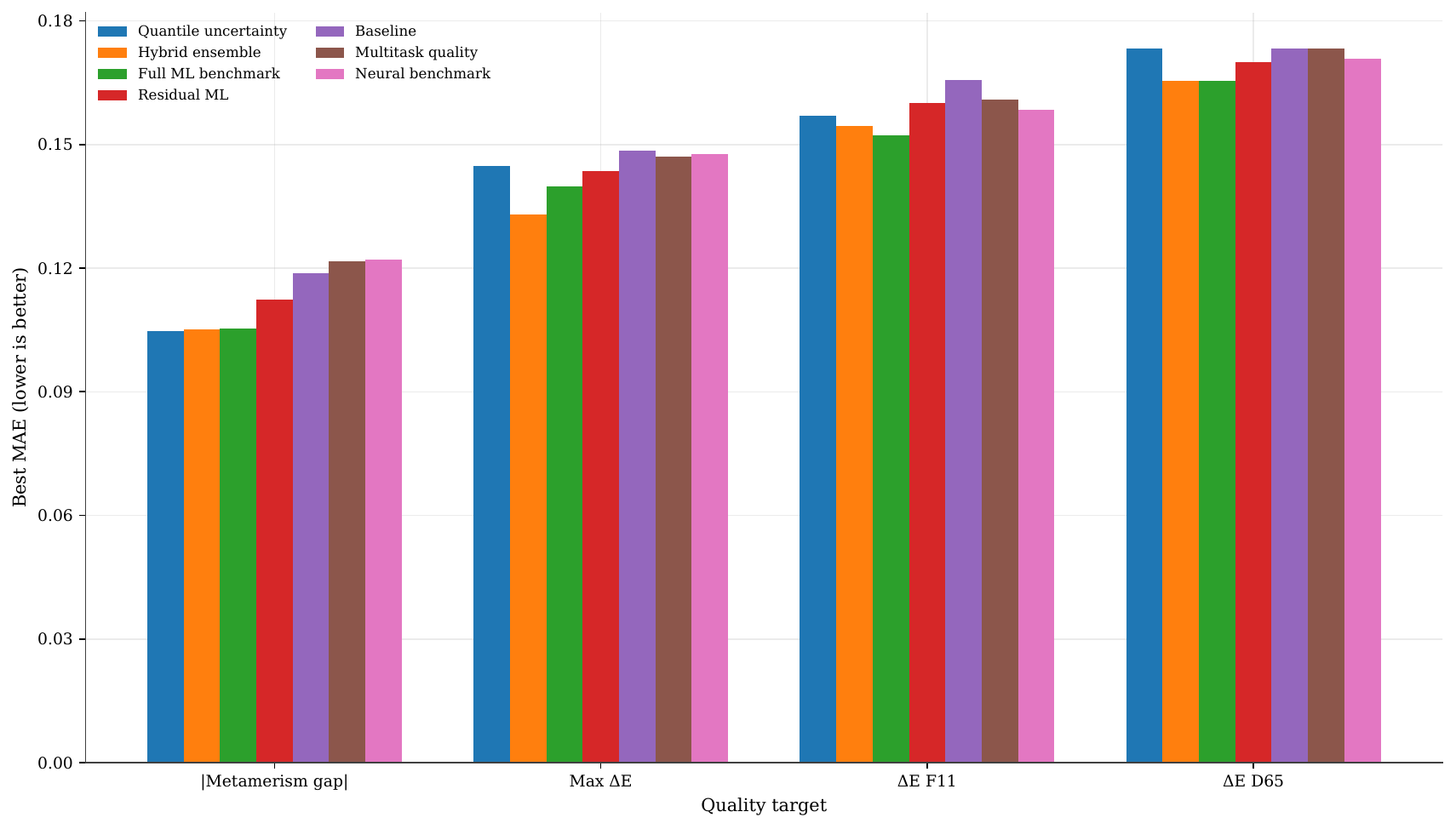}
		\caption{Lowest pooled out-of-fold MAE by model block and color-quality target.}
		\label{fig:lowest_cross_validated_mae}
	\end{figure}
	
	The lowest-MAE model differed across targets, shown in Figures~\ref{fig:forward_observed_vs_predicted} and~\ref{fig:lowest_cross_validated_mae}. Histogram gradient boosting with spectral-only inputs ranked first for \(\Delta E_{\mathrm{D65}}\) and \(\Delta E_{\max}\). Gradient boosting with recipe-process inputs ranked first for \(\Delta E_{\mathrm{F11}}\). And lastly for absolute metamerism gap, a quantile random forest using linear-safe plus K/S features ranked first.
	
	Spearman correlations were lower for \(\Delta E_{\mathrm{D65}}\) (\(\rho=0.2368\)) and \(\Delta E_{\max}\) (\(\rho=0.2507\)) than for \(\Delta E_{\mathrm{F11}}\) (\(\rho=0.3988\)) and the absolute metamerism gap (\(\rho=0.4746\)). As shown in Figure~\ref{fig:forward_absolute_error_distributions}, for every target, median absolute error was lower than MAE. It indicates that larger errors increased the mean. Bias remained close to zero across all four targets.
	
	\subsubsection{Multiplicity-Adjusted Comparison with Reference Baselines}
	Candidate models were compared with the dummy-median predictor and the lowest-MAE model in the baseline-model block for each color-quality target. 
	The multiplicity-adjusted comparisons are shown in Table~\ref{tab:multiplicity_adjusted_claims}. The QRF-LINEAR-KM model exhibits improvement after Holm correction over the fold-specific dummy-median reference for $\lvert\Delta E_{\mathrm{D65}}-\Delta E_{\mathrm{F11}}\rvert$ ($\Delta\mathrm{MAE}=-0.0273$, 95\% CI [$-0.0433$, $-0.0120$], $p_{\mathrm{Holm}}=0.033$) while other models have not appeared as statistically significant.
	
	\begin{table}[H]
		\centering
		\caption{Multiplicity-adjusted comparisons of the selected forward models with reference baselines.}
		\label{tab:multiplicity_adjusted_claims}
		\begin{adjustbox}{max width=\textwidth}
			\begin{tabular}{l l l c c c}
				\toprule
				\textbf{Target}
				& \textbf{Model}
				& \textbf{Reference}
				& \textbf{$\Delta$MAE (95\% CI)}
				& \textbf{Reduction}
				& \textbf{$p_{\mathrm{Holm}}$} \\
				\midrule
				
				$\Delta E_{\mathrm{D65}}$
				& HGB- (Sp.)
				& Median* (best baseline)
				& $-0.0078$ [$-0.0178$, 0.0023]
				& 4.5\%
				& 0.108 \\
				
				\midrule
				
				$\Delta E_{\mathrm{F11}}$
				& GB- (RP)
				& Ridge- (Sp.)
				& $-0.0133$ [$-0.0303$, 0.0028]
				& 8.0\%
				& 0.417 \\
				
				$\Delta E_{\mathrm{F11}}$
				& GB- (RP)
				& Median*
				& $-0.0153$ [$-0.0314$, 0.0008]
				& 9.1\%
				& 0.356 \\
				
				\midrule
				
				$\Delta E_{\max}$
				& HGB- (Sp.)
				& Median* (best baseline)
				& $-0.0087$ [$-0.0196$, 0.0014]
				& 5.9\%
				& 0.417 \\
				
				\midrule
				
				$\lvert\Delta E_{\mathrm{D65}}-\Delta E_{\mathrm{F11}}\rvert$
				& QRF-LINEAR-KM (LSP)
				& Ridge- (Sp.)
				& $-0.0142$ [$-0.0245$, $-0.0043$]
				& 11.9\%
				& 0.085 \\
				
				$\lvert\Delta E_{\mathrm{D65}}-\Delta E_{\mathrm{F11}}\rvert$
				& QRF-LINEAR-KM (LSP)
				& Median*
				& $-0.0273$ [$-0.0433$, $-0.0120$]
				& 20.7\%
				& \textbf{0.033} \\
				
				\bottomrule
			\end{tabular}
		\end{adjustbox}
		
		\vspace{0.2cm}
		
		\begin{minipage}{0.98\linewidth}
			\footnotesize
			\textit{Note.} Median* is the fold-specific dummy-median model.
			$\Delta$MAE uses 212 pooled out-of-fold observations; inference
			uses 85 recipe groups. Negative values favour the selected model.
			Sp.: spectral-only; RP: recipe--process; LSP: linear-safe plus K/S.
		\end{minipage}
	\end{table}
	
	\subsubsection{Input-Block Ablation Across Models-Targets and Feature Importance}
\begin{figure}[H]
		\centering
		\includegraphics[width=\textwidth]
		{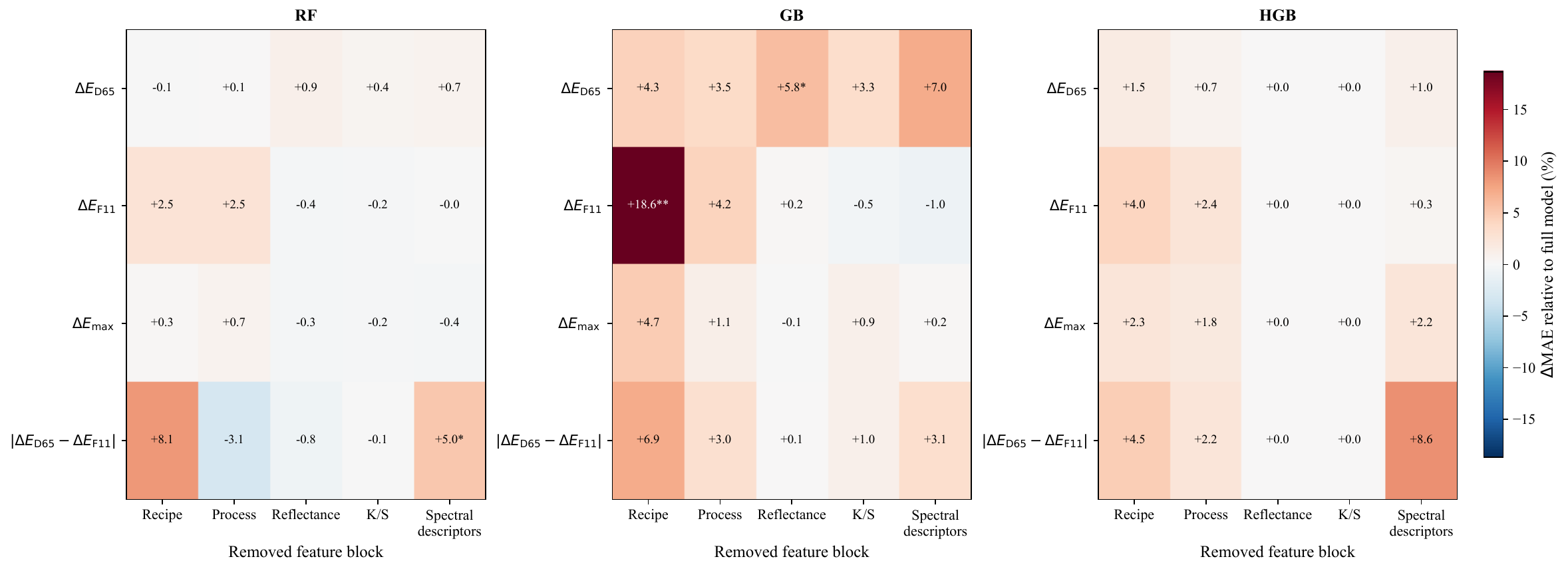}
		\caption{Relative changes in MAE after input-block removal.}
		\label{fig:all_targets_models_ablation_heatmap}
		\par\vspace{0.2cm}
		\begin{minipage}{0.96\textwidth}
			\footnotesize
			\textit{Note.} \textbf{*} denotes significance after within-model--target Holm correction. \textbf{**} indicates significance after both within-family and global Holm correction.
		\end{minipage}
	\end{figure}

\begin{figure}[H]
		\centering
		\includegraphics[width=\textwidth]
		{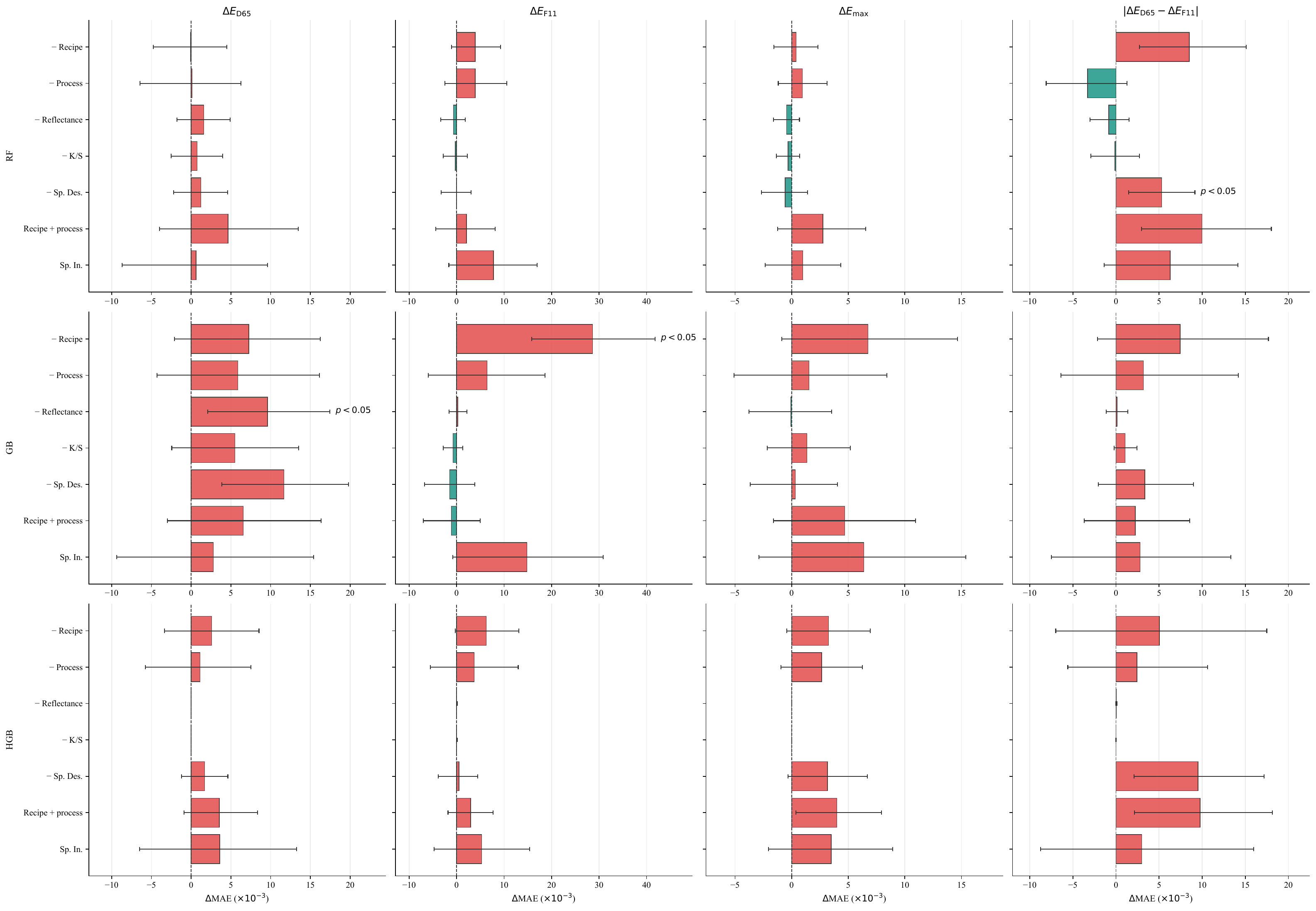}
		\caption{Input-block ablation across models and color-quality targets.}
		\label{fig:all_targets_models_ablation_grid}
	\end{figure}

	Ablation tests across three models (RF, GB, and HGB) and all four targets (\(\Delta E_{\mathrm{D65}}\), \(\Delta E_{\mathrm{F11}}\), \(\Delta E_{\max}\), and \(\lvert\Delta E_{\mathrm{D65}}-\Delta E_{\mathrm{F11}}\rvert\)) are shown in Table~\ref{tab:significant_input_ablation} and Figures~\ref{fig:all_targets_models_ablation_grid} and~\ref{fig:all_targets_models_ablation_heatmap}. 
	
	\begin{table}[H]
		\centering
		\caption{Significant increases in MAE following input-block removal.}
		\label{tab:significant_input_ablation}
		\small
		\begin{adjustbox}{max width=\textwidth}
			\begin{tabular}{l l l c c c c}
				\toprule
				\textbf{Model}
				& \textbf{Target}
				& \textbf{Removed block}
				& \textbf{$\Delta$MAE}
				& \textbf{95\% CI}
				& \textbf{Change}
				& \textbf{$p_{\mathrm{Holm}}$} \\
				\midrule
				
				RF
				& $\lvert\Delta E_{\mathrm{D65}}-\Delta E_{\mathrm{F11}}\rvert$
				& Sp. Des.
				& $+0.0053$
				& [0.0015, 0.0092]
				& $+5.04\%$
				& 0.0218 \\
				
				GB
				& $\Delta E_{\mathrm{D65}}$
				& Reflectance
				& $+0.0096$
				& [0.0021, 0.0174]
				& $+5.75\%$
				& 0.0046 \\
				
				GB
				& $\Delta E_{\mathrm{F11}}$
				& Recipe
				& $+0.0286$
				& [0.0159, 0.0418]
				& $+18.65\%$
				& $<0.001$ \\
				
				\bottomrule
			\end{tabular}
		\end{adjustbox}
		
		\vspace{0.2cm}
		
		\begin{minipage}{0.96\textwidth}
			\footnotesize
			\textit{Note.} Positive values indicate higher error. Sp. Des.: spectral descriptors.
		\end{minipage}
	\end{table}
	The analysis removed five blocks, including recipe, process, reflectance, K/S, and spectral descriptors, separately and tested them against the full model. Recipe-process and spectral-input configurations were used as descriptive controls. Under within-model–target Holm correction, three removals increased MAE presented in Table~\ref{tab:significant_input_ablation}, although only the GB-\(\Delta E_{\mathrm{F11}}\) (See Figure~\ref{fig:all_targets_models_ablation_heatmap}) recipe-feature result survived global Holm correction, \(p_{\mathrm{Holm}}=0.0027\).
	
	RF was quite insensitive to change, as illustrated in Figure~\ref{fig:all_targets_models_ablation_grid}, while GB showed target-specific dependency. None of the HGB ablation was statistically significant. However, \textit{non-significant} refers to \textit{no detectable change}, rather than \textit{not important}. Some feature blocks may contain overlapping or nonlinear dependencies, which might account for these limited effects. 
	
\begin{figure}[H]
		\centering
		\includegraphics[width=\textwidth]
		{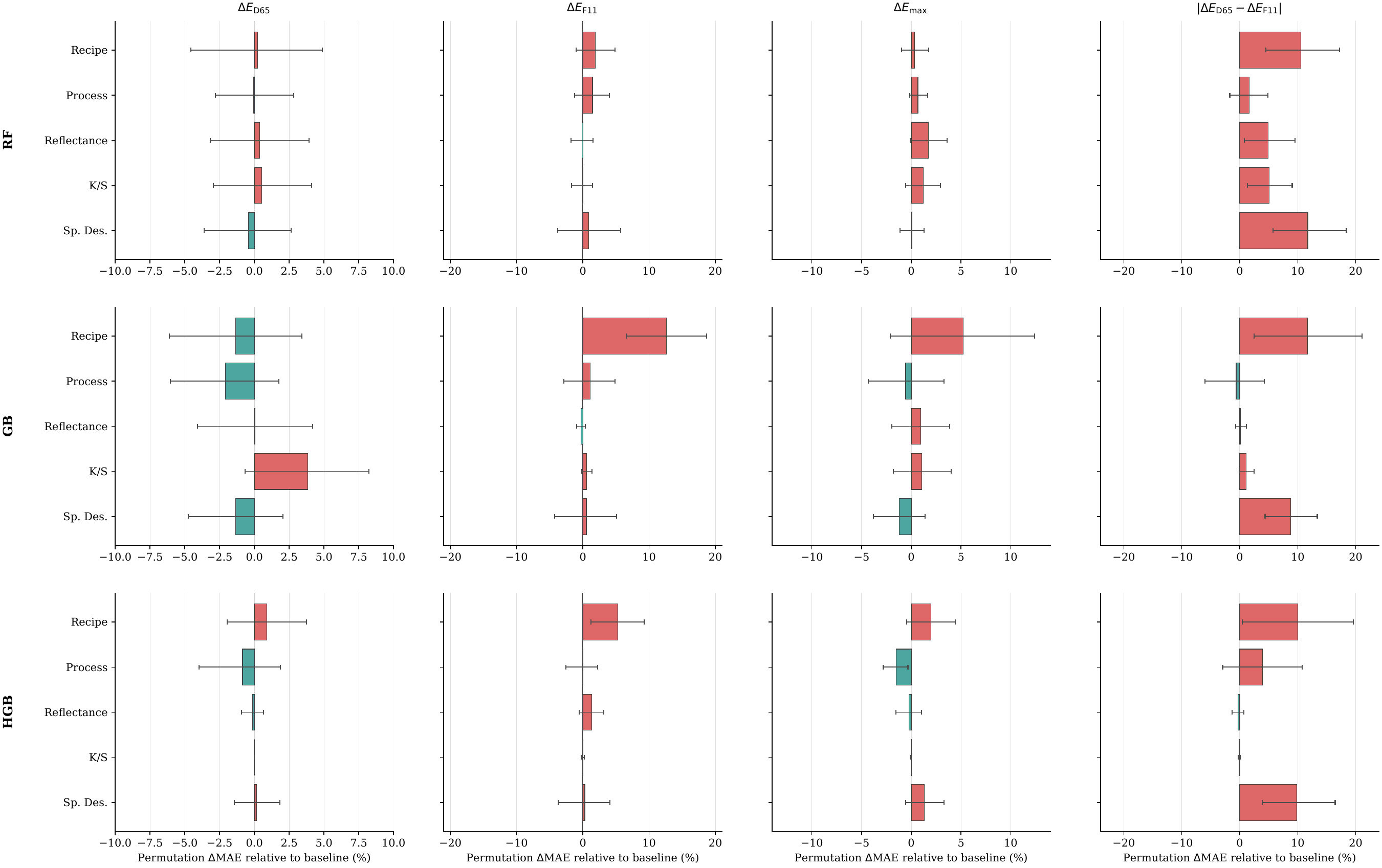}
		\caption{Feature-block permutation importance across models and targets.}
		\label{fig:block_permutation_importance}
		
		\par\vspace{0.2cm}
		
		\begin{minipage}{0.96\textwidth}
			\footnotesize
			\textit{Note.} Positive values indicate increased held-out MAE after permutation. Error bars show 95\% recipe-cluster bootstrap confidence intervals based on 20 joint permutations of each feature block.
		\end{minipage}
	\end{figure}

	The permutation importance, illustrated as Figure~\ref{fig:block_permutation_importance}, measured the increase in held-out MAE after permuting an input block. A positive MAE points out the dependency of that feature in that fitted model. Predicting metamerism exhibits the strongest feature block reliance. For instance, MAE increased for RF (spectral descriptors), GB (recipe features), and HGB (process features) for predicting the absolute metamerism gap of \(+16.27\%\) (95\% CI: \(9.48\)–\(23.69\%\)), \(+16.59\%\) (\(7.77\)–\(25.69\%\)), and \(+16.42\%\) (\(8.10\)–\(24.65\%\)), respectively, which agrees with the ablation finding demonstrated in Figure~\ref{fig:all_targets_models_ablation_grid}.
	
\begin{figure}[H]
		\centering
		\includegraphics[width=\textwidth]
		{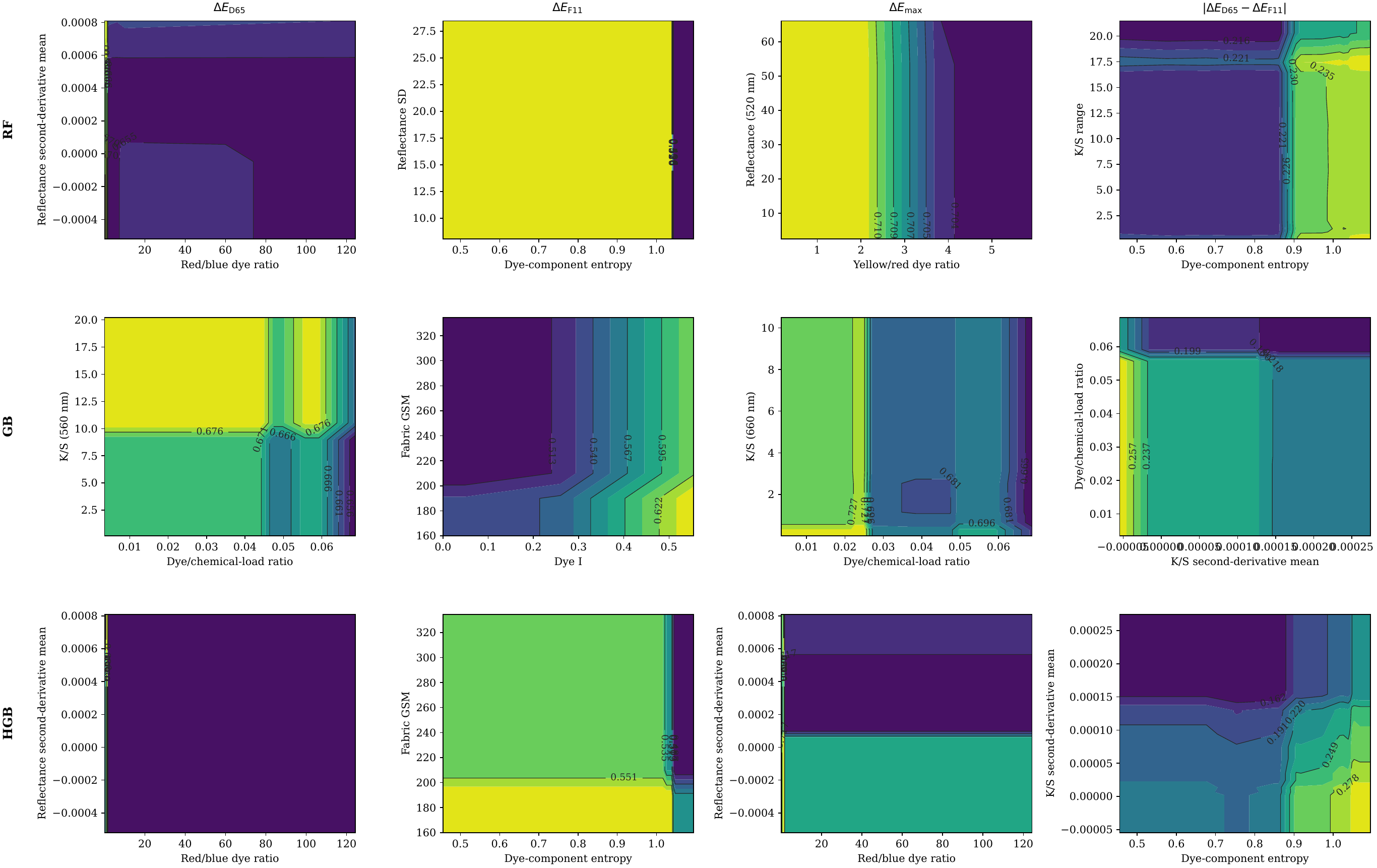}
		\caption{Two-way partial-dependence surfaces across models and targets.}
		\label{fig:two_way_pdp}
		
		\par\vspace{0.2cm}
		
		\begin{minipage}{0.96\textwidth}
			\footnotesize
			\textit{Note.} Contours show mean model predictions over pairs of influential continuous predictors. The surfaces are descriptive and do not represent causal effects.
		\end{minipage}
	\end{figure}

	The two-way partial-dependence surfaces, shown in Figure~\ref{fig:two_way_pdp}, showed the nonlinear and model-specific prediction patterns of features. RF and HGB show lower predictions as the 'red/blue ratio' increases, whereas the increment of dye/chemical-load ratio decreases it, considering \(\Delta E_{\mathrm{D65}}\) as the response variable. In case of predicting \(\Delta E_{\mathrm{F11}}\), an elevated projection was observed for GB for a certain GSM with the addition of ‘Dye I'. However, mostly HGB and RF partial-dependence surfaces are flat. The strongest variation has occurred for the metamerism gap, especially from dye-component entropy and engineered K/S features.
	
	\subsubsection{Prediction-Interval Coverage and Uncertainty Informativeness}
\begin{figure}[H]
		\centering
		\includegraphics[width=0.82\textwidth]{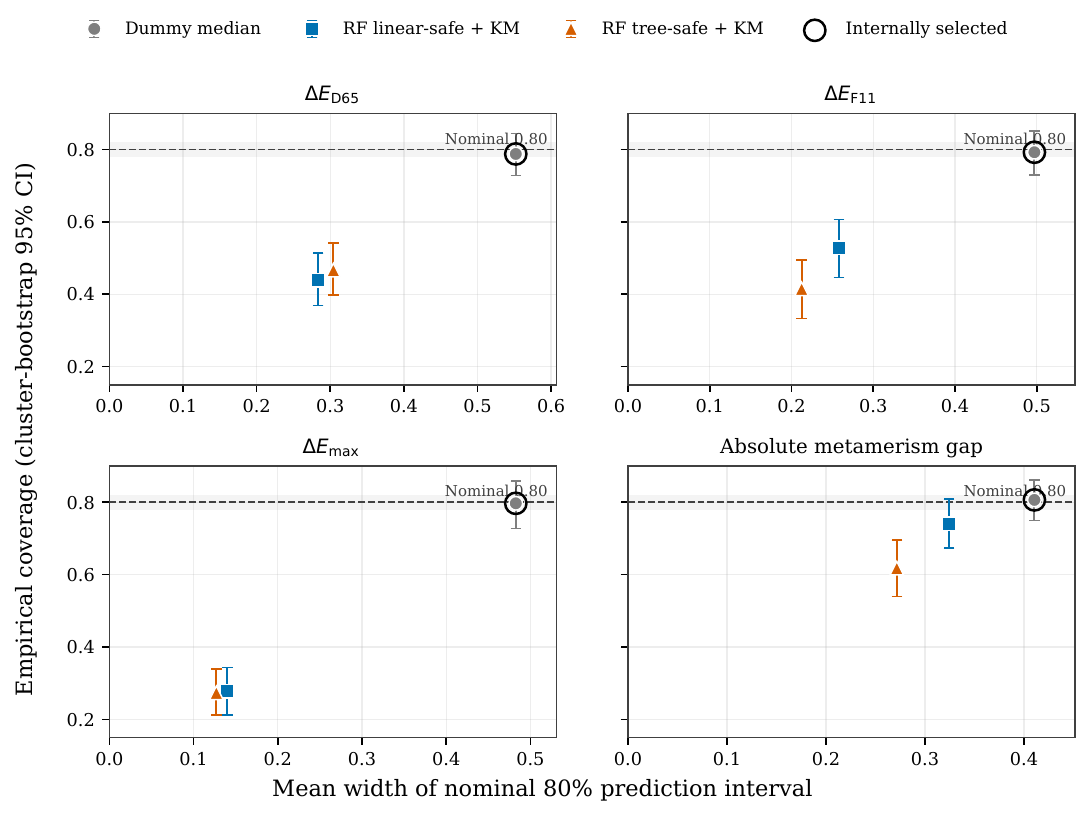}
		\caption{Empirical coverage of nominal 80\% prediction intervals.}
		\label{fig:uncertainty_empirical_coverage}
	\end{figure}

\begin{figure}[H]
		\centering
		\includegraphics[width=0.90\textwidth]{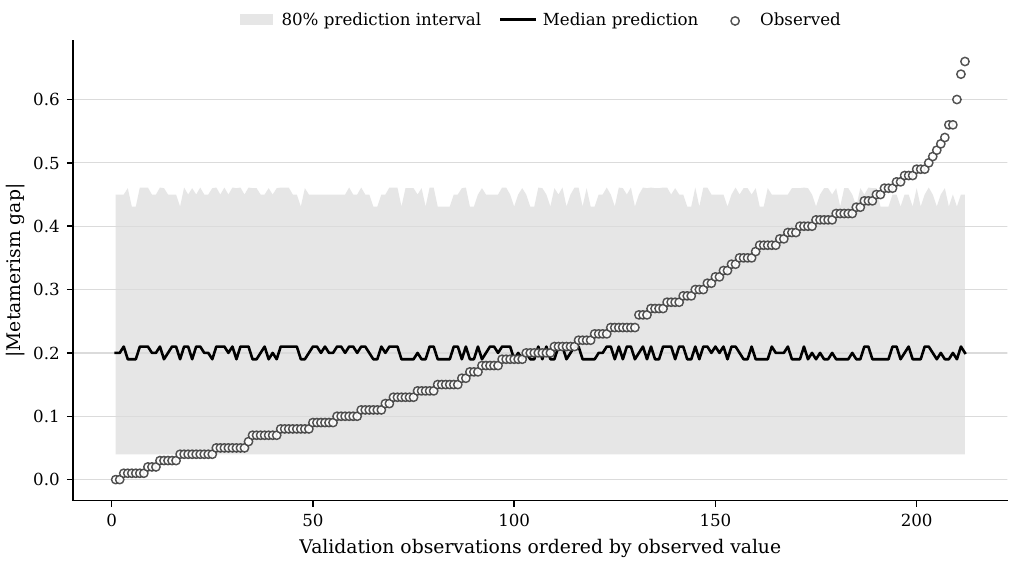}
		\caption{Observed absolute metamerism gaps, median predictions, and nominal 80\% prediction intervals for the selected uncertainty method.}
		\label{fig:prediction_interval_coverage}
	\end{figure}

	The results of uncertainty analysis have been presented in Table~\ref{tab:selected_uncertainty_methods} and Figures~\ref{fig:uncertainty_empirical_coverage} and~\ref{fig:prediction_interval_coverage}. The analysis covered the comparison of three separate methods: fold-wise empirical-quantile baseline, QRF linear-safe + K/S, and QRF tree-safe + K/S. QRF for \(\Delta E_{\mathrm{D65}}\), \(\Delta E_{\mathrm{F11}}\), and \(\Delta E_{\max}\) produces approximately half of the baseline coverage, but their narrow estimate missed apparently half to three-fourths of the observations. However, linear safe QRF offers the best trade-off among the learned baselines with coverage of 0.741 and width 0.324, whose coverage CI included 0.80 alongside better median MAE than the baseline.
	
	\begin{table}[H]
		\centering
		\caption{Prediction-interval performance of the selected fold-wise empirical-quantile method.}
		\label{tab:selected_uncertainty_methods}
		\small
		\begin{adjustbox}{max width=\textwidth}
			\begin{tabular}{l c c c c c}
				\toprule
				\textbf{Target}
				& \textbf{Median MAE (95\% CI)}
				& \textbf{Coverage (95\% CI)}
				& \textbf{Mean width}
				& \textbf{Mean pinball loss}
				& \textbf{Width--error $\rho$} \\
				\midrule
				
				$\Delta E_{\mathrm{D65}}$
				& 0.1733 [0.1546, 0.1933]
				& 0.7877 [0.7277, 0.8436]
				& 0.5520
				& 0.0517
				& 0.0512 \\
				
				$\Delta E_{\mathrm{F11}}$
				& 0.1676 [0.1451, 0.1925]
				& 0.7925 [0.7297, 0.8512]
				& 0.4971
				& 0.0522
				& $-0.0600$ \\
				
				$\Delta E_{\max}$
				& 0.1485 [0.1311, 0.1687]
				& 0.7972 [0.7281, 0.8585]
				& 0.4826
				& 0.0455
				& $-0.0678$ \\
				
				$\lvert\Delta E_{\mathrm{D65}}
				-\Delta E_{\mathrm{F11}}\rvert$
				& 0.1319 [0.1162, 0.1477]
				& 0.8066 [0.7500, 0.8616]
				& 0.4104
				& 0.0387
				& $-0.0414$ \\
				
				\bottomrule
			\end{tabular}
		\end{adjustbox}
		
		\vspace{0.2cm}
		
		\begin{minipage}{0.98\textwidth}
			\footnotesize
			\textit{Note.} Nominal coverage was 0.80; CIs used 2,000 recipe-cluster bootstrap resamples across 85 recipe groups (\(n=212\)). Median MAE refers to median-prediction MAE.
		\end{minipage}
	\end{table}

	For the quantile baseline, illustrated in Table~\ref{tab:selected_uncertainty_methods}, coverage was nearly 0.8 for all four targets with broad mean widths, but width-error correlations were close to zero \((-0.0678\leq\rho\leq0.0512)\). Hence, a larger interval width doesn’t establish a larger prediction error. In addition, the median prediction for the absolute metamerism gap was approximately $0.13$ with a constant interval width ($0.41$), demonstrated at Figure~\ref{fig:prediction_interval_coverage}. Most of the central observations fall within this band. Though QRF produced higher accuracy except for \(\Delta E_{\mathrm{D65}}\), its general trend concluded in substantial undercoverage.
	
	\subsection{Module B: Sparse Inverse Recipe Reconstruction and Meta-Model Generalization}
	\label{sec:results_inverse_reconstruction}
	
	\subsubsection{Component Base Models and Component-Wise Median Blend Meta-Model}
	The result of the inverse recipe reconstruction, illustrated as Table~\ref{tab:inverse_recipe_reconstruction}, points out that median blend, KNN, and extra trees appear top competitor models under out-of-fold test predictions of recipe vectors. The median blend shows the lowest recipe-vector MAE (\(0.0295\)), versus \(0.0327\) for KNN and \(0.0338\) for Extra Trees. It also obtains the highest precision (\(0.7075\)) and F1 (\(0.6887\)) at the same time. Nevertheless, Extra Trees achieves high recall ($0.6942$) with the best composite score among individual models ($0.3805$), while another variant ET-(TSP) has  the lowest total-dye error (\(0.2723\)).
	
\begin{figure}[H]
		\centering
		\includegraphics[width=\textwidth]{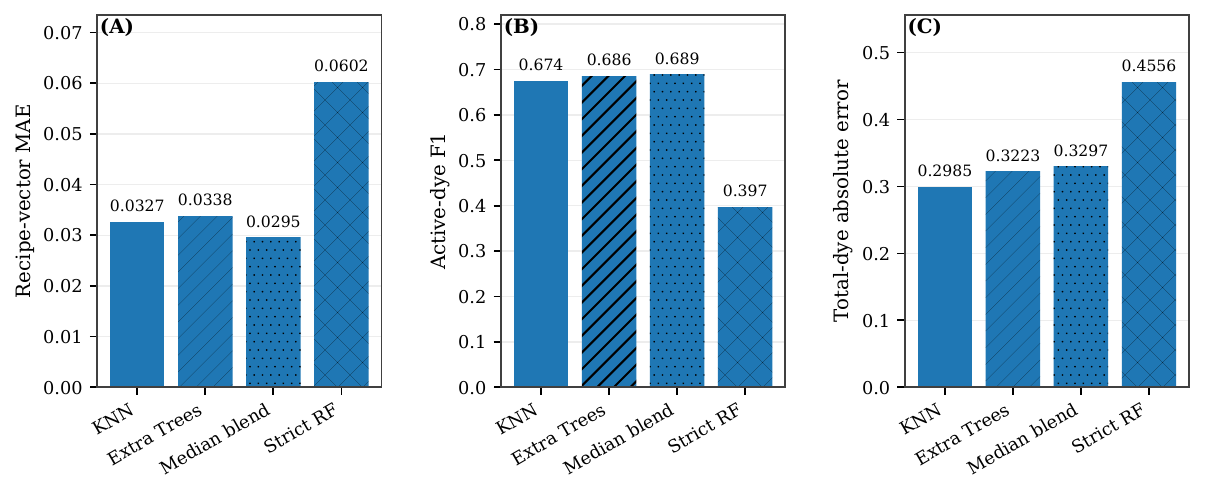}
		\caption{Comparison of selected inverse-reconstruction models using recipe-vector MAE, active-dye F1 score, and total-dye absolute error.}
		\label{fig:inverse_selected_model_metrics}
	\end{figure}

	\begin{table}[H]
		\centering
		\caption{Comparison of inverse base and meta-models under recipe-grouped and strict unseen-target validation.}
		\label{tab:inverse_recipe_reconstruction}
		\scriptsize
		\setlength{\tabcolsep}{3pt}
		\renewcommand{\arraystretch}{1.08}
		
		\begin{adjustbox}{max width=\textwidth}
			\begin{tabular}{l c c c c c c c}
				\toprule
				\textbf{Model}
				& \begin{tabular}[c]{@{}c@{}}\textbf{Recipe-vector}\\\textbf{MAE}\end{tabular}
				& \begin{tabular}[c]{@{}c@{}}\textbf{Total-dye}\\\textbf{error}\end{tabular}
				& \textbf{Precision}
				& \textbf{Recall}
				& \textbf{F1}
				& \begin{tabular}[c]{@{}c@{}}\textbf{Active Top-$k$}\\\textbf{recall}\end{tabular}
				& \begin{tabular}[c]{@{}c@{}}\textbf{Composite}\\\textbf{score}\end{tabular} \\
				\midrule
				
				\multicolumn{8}{l}{\textit{Recipe-grouped validation: base models}} \\
				
				ET- (TS), best individual base
				& 0.0338 & 0.3223 & 0.6824 & 0.6942 & 0.6855 & 0.6892 & 0.3805 \\
				
				ET- (TSP)
				& 0.0319 & 0.2723 & 0.6682 & 0.6965 & 0.6777 & 0.6989 & 0.3814 \\
				
				RF- (TS)
				& 0.0334 & 0.3237 & 0.6745 & 0.6981 & 0.6821 & 0.6787 & 0.3837 \\
				
				KNN- (TS), reference
				& 0.0327 & 0.2985 & 0.6722 & 0.6816 & 0.6736 & 0.6908 & 0.3889 \\
				
				RF- (TSP)
				& 0.0341 & 0.2862 & 0.6612 & 0.6895 & 0.6711 & 0.6626 & 0.3916 \\
				
				KNN- (TSP)
				& 0.0310 & 0.3108 & 0.6675 & 0.6769 & 0.6682 & 0.6932 & 0.3938 \\
				
				ElasticNet- (TS)
				& 0.0526 & 0.6330 & 0.4717 & 0.4741 & 0.4726 & 0.4855 & 0.6433 \\
				
				Ridge- (TS)
				& 0.0549 & 0.6876 & 0.4654 & 0.4678 & 0.4664 & 0.4791 & 0.6573 \\
				
				ElasticNet- (TSP)
				& 0.0614 & 0.4468 & 0.3947 & 0.3947 & 0.3947 & 0.4042 & 0.7114 \\
				
				Ridge- (TSP)
				& 0.0649 & 0.6138 & 0.4057 & 0.4080 & 0.4066 & 0.4179 & 0.7196 \\
				
				\midrule
				
				\multicolumn{8}{l}{\textit{Recipe-grouped validation: meta-models}} \\
				
				Median blend, selected meta-model
				& 0.0295 & 0.3297 & 0.7075 & 0.6808 & 0.6887 & 0.6932 & 0.3737 \\
				
				ElasticNet meta-model
				& 0.0382 & 0.4809 & 0.5519 & 0.5715 & 0.5583 & 0.5612 & 0.5280 \\
				
				Arithmetic-mean blend
				& 0.0402 & 0.3582 & 0.5252 & 0.5401 & 0.5297 & 0.5362 & 0.5463 \\
				
				Ridge meta-model
				& 0.0399 & 0.5124 & 0.5362 & 0.5432 & 0.5390 & 0.5491 & 0.5521 \\
				
				\midrule
				
				\multicolumn{8}{l}{\textit{Strict unseen-target validation: base models}} \\
				
				RF- (TSP), selected strict
				& 0.0602 & 0.4556 & 0.3931 & 0.4025 & 0.3969 & 0.4122 & 0.7089 \\
				
				ET- (TS)
				& 0.0629 & 0.4900 & 0.3954 & 0.4096 & 0.4009 & 0.4147 & 0.7109 \\
				
				RF- (TS)
				& 0.0613 & 0.4703 & 0.3664 & 0.3758 & 0.3701 & 0.3849 & 0.7382 \\
				
				ET- (TSP)
				& 0.0652 & 0.4387 & 0.3522 & 0.3616 & 0.3560 & 0.3704 & 0.7531 \\
				
				KNN- (TSP)
				& 0.0633 & 0.4344 & 0.3318 & 0.3349 & 0.3330 & 0.3430 & 0.7738 \\
				
				KNN- (TS)
				& 0.0717 & 0.6116 & 0.3082 & 0.3153 & 0.3110 & 0.3156 & 0.8219 \\
				
				ElasticNet- (TSP)
				& 0.0771 & 0.5453 & 0.3082 & 0.3082 & 0.3082 & 0.3156 & 0.8235 \\
				
				Ridge- (TSP)
				& 0.0857 & 0.7711 & 0.3035 & 0.3058 & 0.3044 & 0.3060 & 0.8584 \\
				
				ElasticNet- (TS)
				& 0.0938 & 0.9794 & 0.2657 & 0.2681 & 0.2667 & 0.2746 & 0.9250 \\
				
				Ridge- (TS)
				& 0.0984 & 1.1102 & 0.2720 & 0.2744 & 0.2730 & 0.2810 & 0.9364 \\
				
				\bottomrule
			\end{tabular}
		\end{adjustbox}
		
		\vspace{0.2cm}
		
		\begin{minipage}{0.98\textwidth}
			\footnotesize
			\textit{Note.} TS: target spectral; TSP: target spectral plus process. Metrics use feasible sparse out-of-fold predictions. Concentration errors are in percentage points (\% OWF).
		
		\end{minipage}
	\end{table}
	However, most of the scores substantially collapsed during strict inverse prediction, as evident from Table~\ref{tab:inverse_recipe_reconstruction} and Figure~\ref{fig:inverse_selected_model_metrics}, where the F1 score decreased to 0.3969. Similarly, recall and precision dropped remarkably, and error metrics demonstrated a rapid growth. This result denotes that although tree-based models can predict near-shade recipes reasonably well, these models' performance degrades when predicting totally unknown recipes.
	
	\subsubsection{Total Dye Concentration Agreement}
	
\begin{figure}[H]
		\centering
		\includegraphics[width=\textwidth]{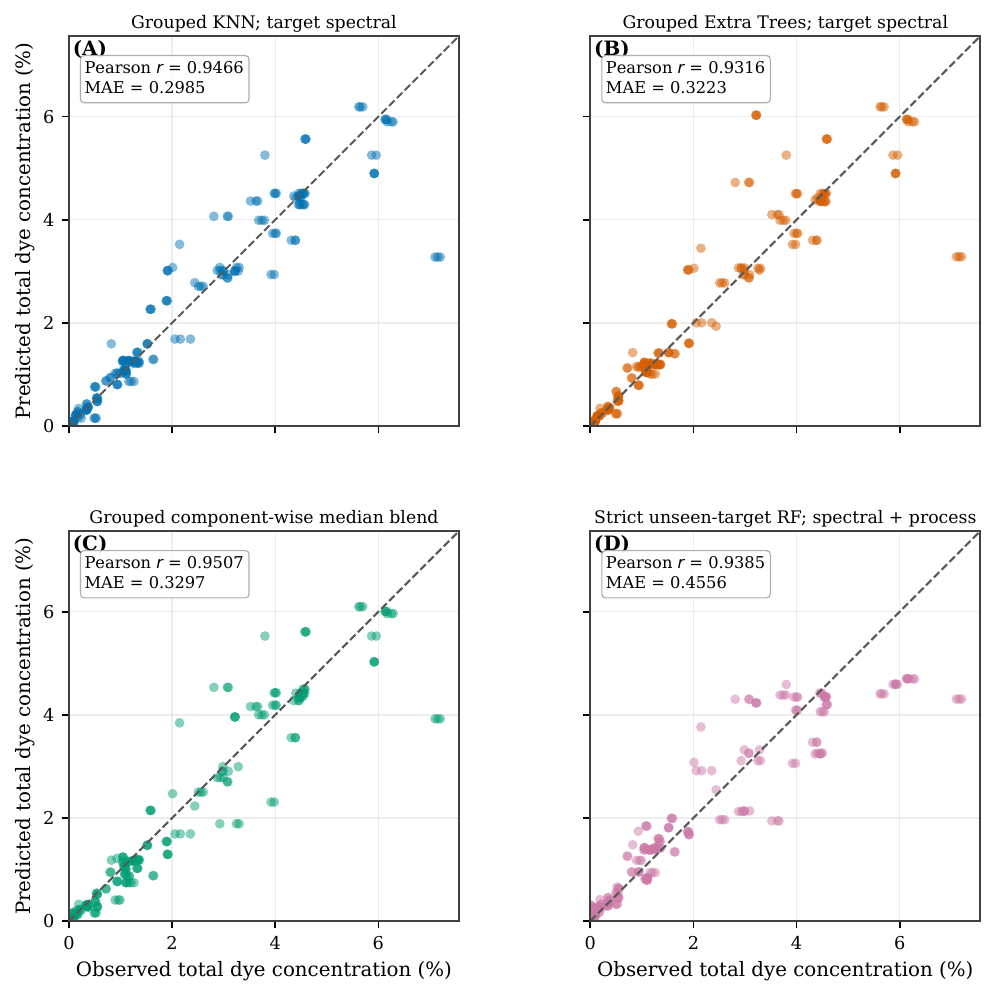}
		\caption{Observed and predicted total dye concentrations for the KNN reference, selected Extra Trees base model, component-wise median blend, and strict unseen-target random forest.}
		\label{fig:inverse_total_dye_all_models}
	\end{figure}

	Observed-predicted correlation (\(r=0.932\)-\(0.951\)), presented as Figure~\ref{fig:inverse_total_dye_all_models} for three grouped models and a strict unseen model, exhibits higher correlation. The correlation is highest for 'median blend,' albeit among the four representative models plotted in Figure~\ref{fig:inverse_total_dye_all_models}, the lowest total-dye error ($0.2985$) was exhibited by the KNN reference (compared to $0.2723$ for ET-(TSP) overall). Figure~\ref{fig:inverse_total_dye_all_models} also suggests that errors have been propagated for several high-load recipes for all four models. This error growth is more visible for strict unseen predictions; even some extreme recipes move aside from the identity line rather than spread and grow symmetrically like the other three.
	
	\subsection{Module C: Feasibility-Constrained Retrieval, Candidate Fusion, and Auditing}
	\label{sec:results_retrieval_fusion}
	
	\subsubsection{Historical Recipe Retrieval Performance}
	
	Historical recipe retrieval results are presented in Table~\ref{tab:retrospective_retrieval_performance} and compare recipe process similarity against outcome-based similarity under Top-\(k\) protocols. ``Top-1'' stands for the \(\Delta E_{\max}\) of the first-ranked candidate. Top-3 and Top-5 are the lowest observed \(\Delta E_{\max}\) among the first three and five candidates, respectively.
	
\begin{figure}[H]
		\centering
		\includegraphics[width=0.95\textwidth]
		{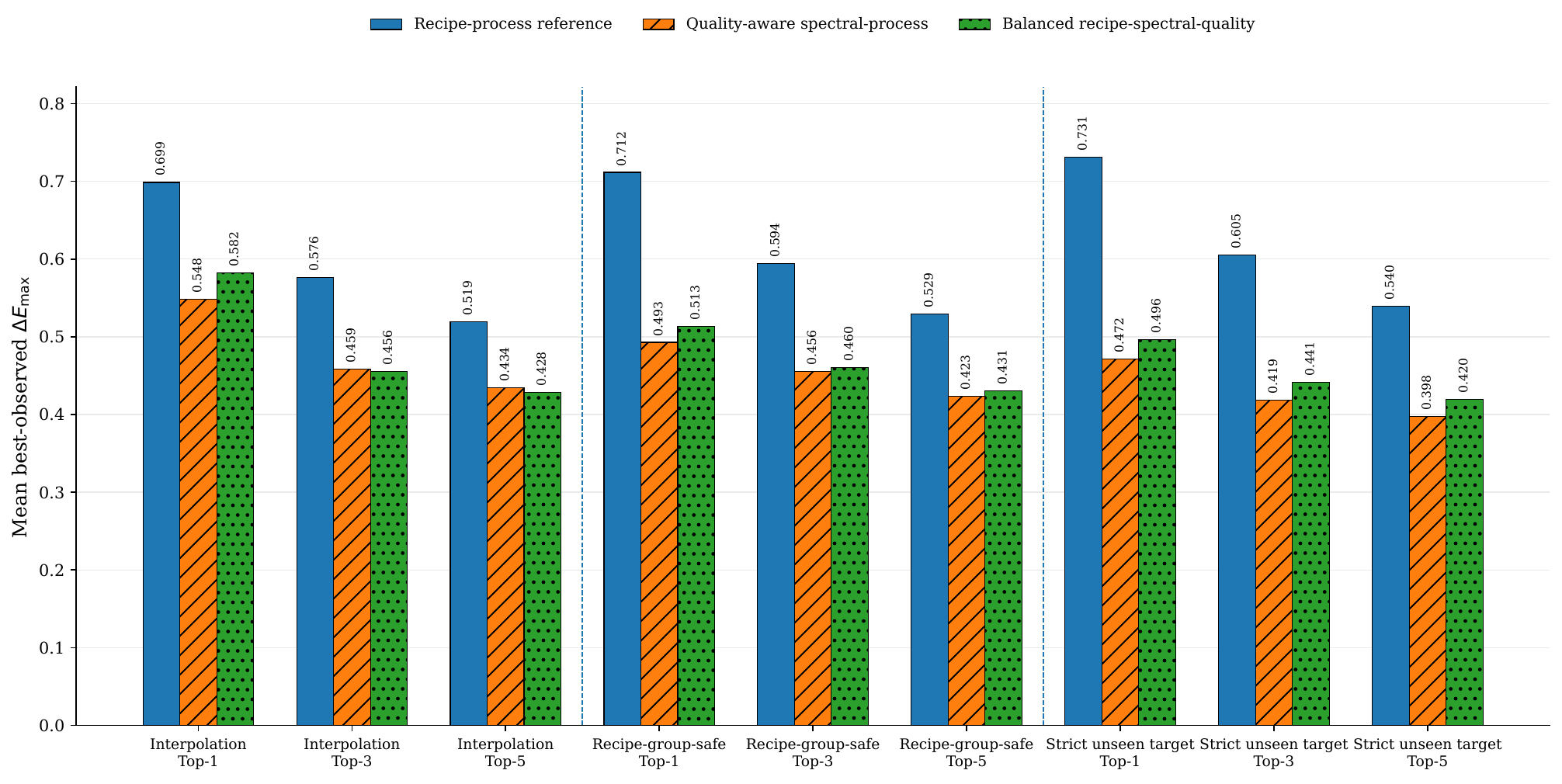}
		\caption{Mean best-observed \(\Delta E_{\max}\) within Top-$k$ historical candidate sets under the three retrieval protocols.}
		\label{fig:top_k_historical_retrieval}
	\end{figure}

	\begin{table}[H]
		\centering
		\caption{Retrospective performance of outcome-informed retrieval.}
		\label{tab:retrospective_retrieval_performance}
		\small
		\begin{adjustbox}{max width=\textwidth}
			\begin{tabular}{l c l c c c}
				\toprule
				\textbf{Protocol}
				& \textbf{Top-$k$}
				& \textbf{Profile}
				& \textbf{Reference mean}
				& \textbf{Profile mean}
				& \textbf{Reduction (95\% CI)} \\
				\midrule
				
				\multirow{3}{*}{Interpolation}
				& 1
				& Sp.
				& 0.6987
				& 0.5482
				& 0.1505 [0.1275, 0.1742] \\
				
				& 3
				& \multirow{2}{*}{Balanced}
				& 0.5765
				& 0.4558
				& 0.1207 [0.1026, 0.1392] \\
				
				& 5
				&
				& 0.5192
				& 0.4282
				& 0.0910 [0.0756, 0.1069] \\
				
				\midrule
				
				\multirow{3}{*}{Recipe-group-safe}
				& 1
				& \multirow{3}{*}{Sp.}
				& 0.7116
				& 0.4928
				& 0.2187 [0.1898, 0.2464] \\
				
				& 3
				&
				& 0.5939
				& 0.4558
				& 0.1381 [0.1111, 0.1644] \\
				
				& 5
				&
				& 0.5293
				& 0.4235
				& 0.1058 [0.0871, 0.1244] \\
				
				\midrule
				
				\multirow{3}{*}{Strict}
				& 1
				& \multirow{3}{*}{Sp.}
				& 0.7315
				& 0.4716
				& 0.2599 [0.2330, 0.2863] \\
				
				& 3
				&
				& 0.6053
				& 0.4187
				& 0.1867 [0.1668, 0.2071] \\
				
				& 5
				&
				& 0.5395
				& 0.3979
				& 0.1417 [0.1236, 0.1589] \\
				
				\bottomrule
			\end{tabular}
		\end{adjustbox}
		
		\vspace{0.2cm}
		
		\begin{minipage}{0.98\textwidth}
			\footnotesize
			\textit{Note.} Sp.: Spectral-process; Reference: recipe--process nearest neighbour; reduction = reference minus profile mean. CIs used 5,000 paired
			query-level bootstrap resamples.
		\end{minipage}
	\end{table}
	Outcome-informed profiles stand out against the recipe process baseline, illustrated in Figure~\ref{fig:top_k_historical_retrieval}, comparing all top-k retrievals. The spectral–process profile had the lowest mean at Top-1, whereas the balanced was marginally better at Top-3 and Top-5. Two global trends in Figure~\ref{fig:top_k_historical_retrieval} may look confusing. Interpolation permits closely related candidates, which is perceptible by the first trend at Top-1: \(0.699 < 0.712 < 0.731\). But for outcome-informed profiles, the second global trend on exhibits is Spectral-process Top-1: \(0.548 \rightarrow 0.493 \rightarrow 0.472\). Nonetheless, it does not prove better generalizability of strict retrieval because of \(\Delta E_{\max}\) reduction. So, the comparison should probably be more plausible between the reference and profile within each protocol, rather than between protocols.
	
	\subsubsection{Cross-Validated Comparison of Historical and Inverse Candidate Sources}
	Four candidate sources were compared using the same fold-specific forward models: historical retrieval, raw and sparse inverse predictions, and the training-fold historical recipe nearest to each prediction.
	
	\paragraph{Recipe Reconstruction Agreement}
	\label{sec:candidate_source_recipe_agreement}
	
\begin{figure}[H]
		\centering
		\includegraphics[width=\textwidth]{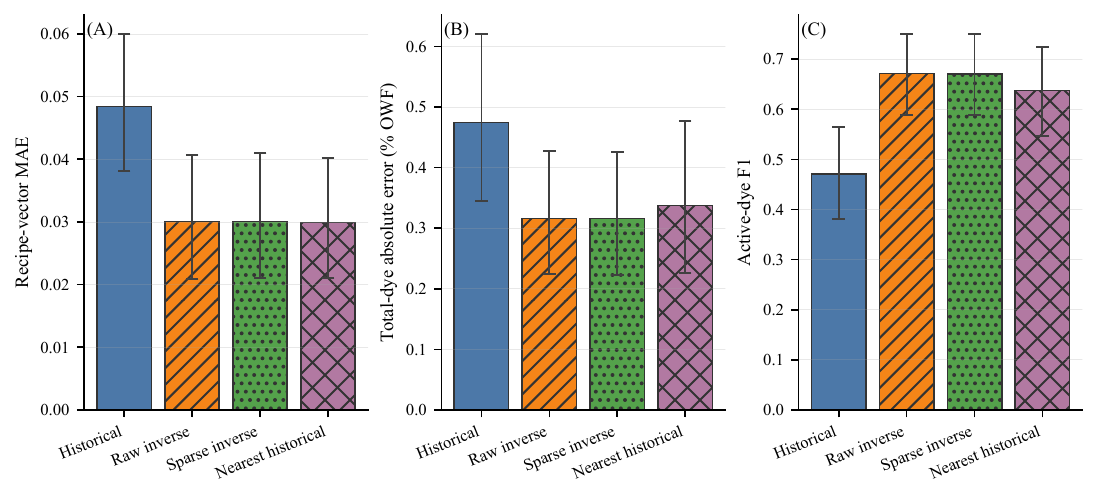}
		\caption{Comparison of candidate sources using recipe-vector MAE, total-dye-load absolute error, and active-dye F1. Error bars show 95\% recipe-cluster bootstrap confidence intervals.}
		\label{fig:candidate_source_recipe_metrics}
	\end{figure}

	The reconstruction agreement with minimum error is one of the focal points of this experiment. Figure~\ref{fig:candidate_source_recipe_metrics} explains and compares the errors made by the constructors. It shows the comparison of historical retrieval, raw and sparse inverse predictions, and the training-fold historical recipe nearest to each prediction against their recipe-vector MAE. The raw and sparse inverse shows lower error rates than historical retrieval with reduced recipe-vector MAE (\(\approx0.030\)), which was \(0.0484\) for historical. Similarly, the active-dye F1 score also increased from \(0.471\) to approximately \(0.671\) during inverse reconstruction. However, upgrading from raw inverse to sparse did decrease F1 and recall slightly (F1: from $0.6717$ to $0.6706$; recall: from $0.6706$ to $0.6627$, which is indeed intuitive), besides increasing precision to some extent, although the change was quite negligible.
	
\begin{figure}[H]
		\centering
		\includegraphics[width=\textwidth]{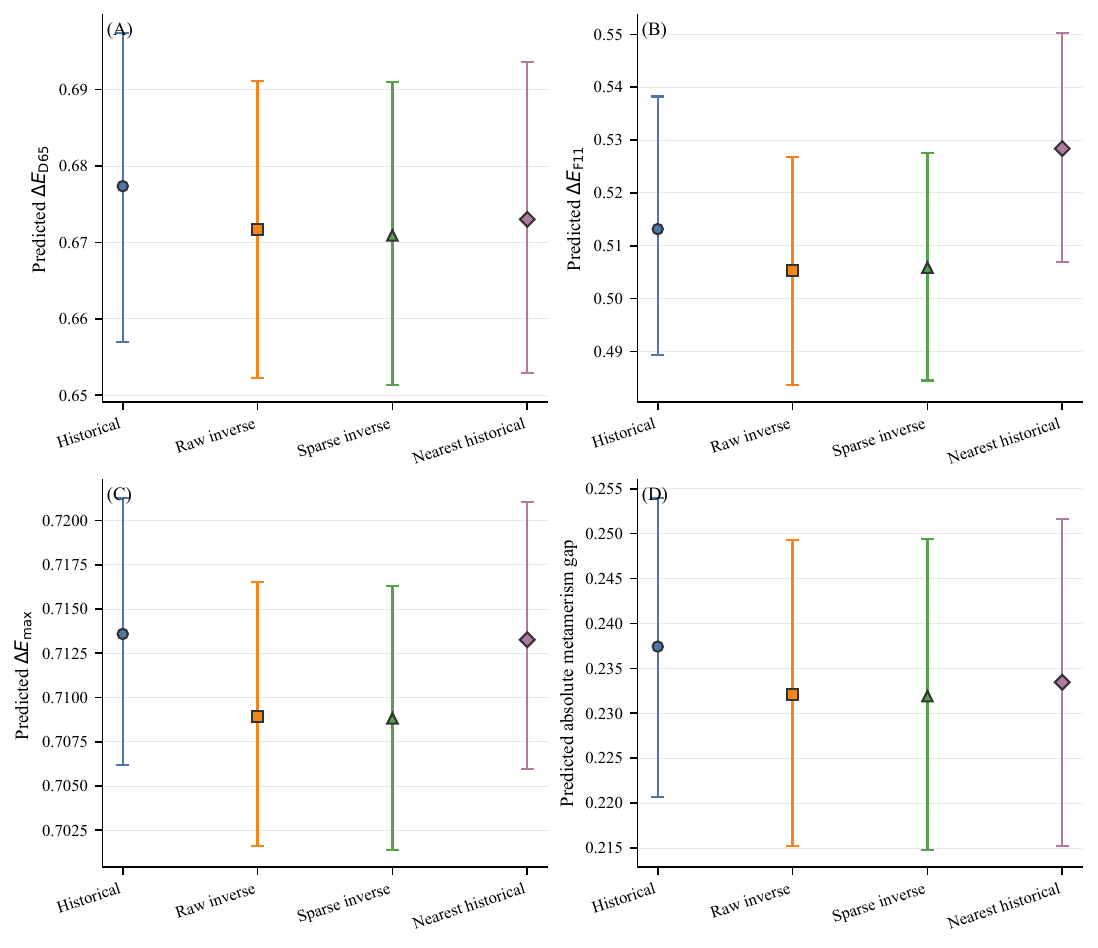}
		\caption{Fold-specific surrogate color-quality predictions for the four candidate sources. Error bars show 95\% recipe-cluster bootstrap confidence intervals.}
		\label{fig:candidate_source_surrogate_quality}
	\end{figure}

	\paragraph{Surrogate Colour-Quality Comparison}
	\label{sec:candidate_source_surrogate_quality}
	The surrogate screening was evaluated using the fold-specific forward models listed in the Table~\ref{tab:inverse_forward_surrogate_mapping} for four candidate sources. Lower predicted \(\Delta E\) refers to promising surrogate color quality. Raw and sparse have negligible result differences. 
	\begin{table}[H]
		\centering
		\caption{Inverse-to-forward model mapping used for surrogate screening.}
		\label{tab:inverse_forward_surrogate_mapping}
		\small
		\begin{tabular}{l l l l}
			\toprule
			\textbf{Inverse model}
			& \textbf{Candidate form}
			& \textbf{Forward outcome}
			& \textbf{Forward model} \\
			\midrule
			
			\multirow{4}{*}{Median blend}
			& \multirow{4}{*}{Raw and sparse}
			& \(\Delta E_{\mathrm{D65}}\)
			& GB- (TSF) \\
			
			&
			& \(\Delta E_{\mathrm{F11}}\)
			& GB- (TSF) \\
			
			&
			& \(\Delta E_{\max}\)
			& RF- (TSF) \\
			
			&
			& \(\lvert\Delta E_{\mathrm{D65}}-\Delta E_{\mathrm{F11}}\rvert\)
			& RF- (TSF) \\
			
			\bottomrule
		\end{tabular}
		
		\vspace{0.2cm}
		
		\begin{minipage}{0.92\textwidth}
			\footnotesize
			\textit{Note.} GB: gradient boosting; RF: random forest; TSF: tree-safe full. The same forward models screened all candidate sources.
		\end{minipage}
	\end{table}
	Though \(\Delta E_{\mathrm{D65}}\) (around 0.67) and metamerism gaps were similar (\(0.232\)-\(0.237\)), they showed similar results across all four sources, illustrated in Figure~\ref{fig:candidate_source_surrogate_quality}. But they appeared globally not significant after the Holm correction. The raw and sparse inverse showed statistically significant improved performance over the nearest historical-to-inverse source for targeting \(\Delta E_{\mathrm{F11}}\) and \(\Delta E_{\max}\), with results after global Holm Correction (for \(\Delta E_{\mathrm{F11}}\), raw: \(p_{\mathrm{Holm}}=0.0159\); sparse: \(p_{\mathrm{Holm}}=0.0187\); for \(\Delta E_{\max}\), raw: \(p_{\mathrm{Holm}}=0.00137\); sparse: \(p_{\mathrm{Holm}}=0.00145\)). Consequently, it can’t be truly concluded that inverse prediction is globally significant across every target over historical retrieval.
	
	\subsection{Exploratory Heterogeneous Treatment-Effect Screening}
	\label{sec:results_causal_ml}
\begin{figure}[H]
		\centering
		\includegraphics[width=\textwidth]{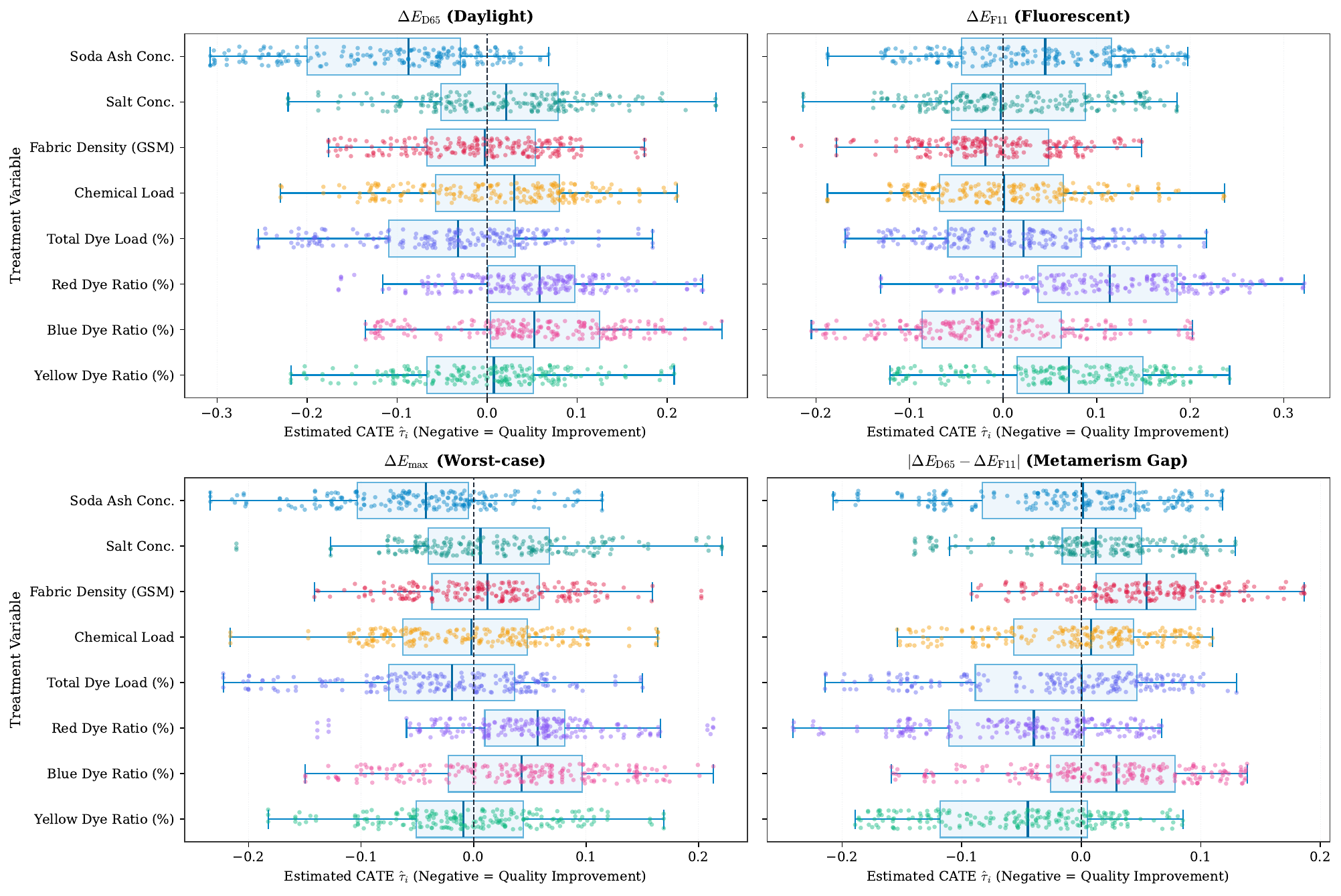}
		\caption{Unit-level CATE distributions ($\hat{\tau}_i$) across 8 process and recipe treatments for the 4 quality targets. Negative values indicate error reduction (quality improvement).}
		\label{fig:cate_distributions}
	\end{figure}

	CATE compares the model-predicted outcome under above-median versus at-or-below-median value, as shown in Table~\ref{tab:cate_heterogeneity}. Figure~\ref{fig:cate_distributions} illustrates the overall distribution on treatment-target combinations. Negative CATE means lower predicted error under high exposure and vice versa. However, these nine subgroup contrasts are in-sample observational RF T-learner estimates.
	
	In the red family and high-reflectance subgroup, higher soda ash corresponded to lower predicted \(\Delta E_{\mathrm{D65}}\). In low reflectance, a higher blue-dye ratio corresponded to a higher predicted \(\Delta E_{\mathrm{D65}}\). In the blue-family subgroup (mid-reflectance), higher red-dye ratio conformed to higher predicted \(\Delta E_{\mathrm{F11}}\). For low-reflectance recipes, high GSM value was associated with a larger predicted metamerism gap. For red-family recipes, a higher yellow-dye ratio was associated with a smaller predicted metamerism gap.
	\begin{table}[H]
		\centering
		\caption{Selected exploratory subgroup CATE estimates.}
		\label{tab:cate_heterogeneity}
		\small
		\begin{tabular}{l l l r}
			\toprule
			\textbf{Target}
			& \textbf{High exposure}
			& \textbf{Subgroup}
			& \textbf{Mean CATE (\(n\))} \\
			\midrule
			
			$\Delta E_{\mathrm{D65}}$
			& Soda ash concentration
			& Red dye family
			& $-0.196$ (62) \\
			
			&
			Soda ash concentration
			& High reflectance
			& $-0.152$ (64) \\
			
			&
			Blue-dye ratio
			& Low reflectance
			& $+0.132$ (71) \\
			
			\midrule
			
			$\Delta E_{\mathrm{F11}}$
			& Red-dye ratio
			& Mid reflectance
			& $+0.159$ (70) \\
			
			&
			Red-dye ratio
			& Blue dye family
			& $+0.135$ (61) \\
			
			\midrule
			
			$\Delta E_{\max}$
			& Soda ash concentration
			& Red dye family
			& $-0.104$ (62) \\
			
			&
			Blue-dye ratio
			& Low reflectance
			& $+0.106$ (71) \\
			
			\midrule
			
			$\lvert\Delta E_{\mathrm{D65}}
			-\Delta E_{\mathrm{F11}}\rvert$
			& Fabric GSM
			& Low reflectance
			& $+0.088$ (71) \\
			
			&
			Yellow-dye ratio
			& Red dye family
			& $-0.083$ (67) \\
			
			\bottomrule
		\end{tabular}
		
		\vspace{0.2cm}
		
		\begin{minipage}{0.90\textwidth}
			\footnotesize
			\textit{Note.} Exploratory observational RF T-learner estimates; negative values indicate lower predicted error under high exposure.
		\end{minipage}
	\end{table}

	\section{Discussion}
	
	This study evaluated a three-module workflow for forward color-quality prediction, sparse inverse reconstruction, and candidate retrieval and screening, together with uncertainty and exploratory interpretation analyses.
	
	\subsection{Forward Prediction and Uncertainty}
	
	Forward performance depended on the prediction target. No model family or feature representation performed best across all four outcomes. Predictions for \(\Delta E_{\mathrm{D65}}\) and \(\Delta E_{\max}\) were concentrated near the centre of the observed range. It shows weaker performance for extreme cases. The selected 80\% prediction intervals achieved near-nominal marginal coverage. However, interval width was weakly related to individual prediction error and varied little within each validation fold. The intervals therefore describe overall uncertainty but cannot reliably identify difficult individual recipes.
	
	\subsection{Historical Retrieval, Inverse Reconstruction, and Candidate Sources}
	
	Outcome-informed retrieval returned historical candidate sets with lower recorded \(\Delta E_{\max}\) than the recipe-process reference. However, candidate quality was used in the ranking score. The result therefore shows retrospective re-ranking rather than prospective performance on untested recipes. Inverse reconstruction produced sparse recipe proposals with reasonable agreement under recipe-grouped validation, but performance declined for unseen target shades. The component-wise median blend gave the lowest recipe-vector MAE, while sparse post-processing retained almost the same performance as the raw inverse predictions. This suggests that limiting the output to three active dyes did not reduce reconstruction accuracy. Raw and sparse inverse candidates matched the recorded recipes better than direct historical retrieval. Replacing a sparse inverse proposal with its nearest historical recipe did not improve reconstruction or surrogate color-quality predictions. These candidates should therefore be treated as candidate for laboratory testing rather than experimentally validated result.
	
	\subsection{Interpretability}
	
	Figure~\ref{fig:block_permutation_importance} reports feature-block permutation importance across models and targets. The metamerism models showed dependence on spectral-descriptor, recipe, and process features. However, the patterns differed across model families. These results are model-specific and do not show causal effects. Correlated predictors can divide or redistribute importance. The ablation results in Table~\ref{tab:significant_input_ablation} and Figures~\ref{fig:all_targets_models_ablation_grid} and~\ref{fig:all_targets_models_ablation_heatmap} identified three significant effects after within-model-target Holm correction. Only the removal of recipe features from GB for \(\Delta E_{\mathrm{F11}}\) remained significant after global Holm correction. Nonsignificant results do not prove that the corresponding feature blocks had no predictive value.
	
	\subsection{Expert Decision-Support System and Practical Implementation}
	\label{sec:discussion_decision_support}
	Rather than acting as an autonomous black-box replacement for laboratory colorists, the proposed hybrid framework is designed as an Expert Decision-Support System (EDSS). In practical mill deployment, the workflow proceeds in three human-in-the-loop stages:
	\begin{enumerate}[label=(\roman*), leftmargin=*]
		\item \textbf{Candidate Generation}: The system retrieves top historical matches and generates sparse inverse formulations ($q \le 3$ active dyes).
		\item \textbf{Surrogate Auditing and Uncertainty Assessment}: Forward models predict multi-illuminant color differences ($\Delta E_{\mathrm{D65}}, \Delta E_{\mathrm{F11}}, \Delta E_{\max}$) alongside 80\% prediction intervals, providing an indication of marginal prediction uncertainty prior to chemical dispensing.
		\item \textbf{Process-Tuning Hypotheses}: Colorists leverage the CATE subgroup insights (e.g., increasing soda ash for red-family or high-reflectance shades, while avoiding dye oversaturation in dark shades) to guide subsequent laboratory testing initial physical lab dips.
	\end{enumerate}
	Although this framework has been applied on textile color matching, the three-module architecture (surrogate quality prediction, sparse inverse reconstruction, and candidate ranking) generalizes to other sparse formulation domains such as industrial coatings, polymer compounding, and chemical product design.
	
	\subsection{Limitations and Future Work}
	
	The dataset contained only 85 recipe-number groups from one industrial source. Generalisation to new substrates, dye systems, process conditions, and production environments is therefore uncertain. Hyperparameter selection and performance estimation used the same grouped folds, and no external or wet-lab validation was conducted. The inverse problem is also non-unique, while sparse post-processing did not evaluate chemical compatibility, concentration limits, cost, or production feasibility. Prediction intervals were mainly calibrated at the aggregate level, and surrogate screening measured agreement between fitted models rather than physical recipe performance. The study also did not directly measure reductions in laboratory trials, development time, or resource use. Future work should evaluate the framework using independent production data and controlled dyeing trials. Inverse models should generate several candidate recipes under chemical, process, concentration, cost, and color-quality constraints. The CATE estimates were exploratory; therefore, they should not be interpreted as confirmed intervention effects. Larger datasets, nested validation, and improved uncertainty methods are also needed before industrial integration. 
	
	\section{Conclusion}
	\label{sec:conclusion}
	
	This study developed and evaluated a physics-informed hybrid machine-learning framework for sparse recipe formulation and applies it to textile color reconstruction. Different models with different types of configurations were tested and some of them showed their superiority based on different targets. Histogram gradient boosting with spectral-only inputs obtained MAEs of \(0.1655\) for \(\Delta E_{\mathrm{D65}}\) and \(0.1398\) for \(\Delta E_{\max}\). Gradient boosting with recipe-process inputs achieved an MAE of \(0.1523\) for \(\Delta E_{\mathrm{F11}}\), while a quantile random forest achieved an MAE of \(0.1046\) for the absolute metamerism gap. After multiplicity adjustment, only the selected metamerism-gap model showed a detectable improvement. For inverse recipe reconstruction, a component-wise median ensemble of ten trained base models achieved the lowest recipe-vector MAE of \(0.0295\). Among the individual models, Extra Trees with spectral inputs achieved a recipe-vector MAE of \(0.0338\). But inverse performance declined for the unseen target shade, with MAE increasing to \(0.0602\). There was no optimal model found in this study that could outperform all tasks simultaneously at a time. However, this framework can be used as a supporting tool rather than an entirely automatic recipe formulation mechanism due to its current limitation of wet-lab-based validation. Future studies can explore more robust dataset with wet-lab validation with industrial deployment target.
	
	\newpage
	\setcounter{page}{1}
	\renewcommand{\thepage}{R-\arabic{page}}
	
	\bibliography{references}
	
	\newpage
	\setcounter{page}{1}
	\renewcommand{\thepage}{A-\arabic{page}}
	\renewcommand{\thetable}{A\arabic{table}}
	\appendix
	\setcounter{table}{0}
	
	\section*{Appendix A: Supplementary Methods and Tables}
	\addcontentsline{toc}{section}{Appendix A: Supplementary Tables}
	
	\begin{table}[H]
		\centering
		\caption{Hyperparameter search spaces used for forward color-quality prediction.}
		\label{tab:forward_hyperparameter_search}
		\scriptsize
		\setlength{\tabcolsep}{4pt}
		\renewcommand{\arraystretch}{1.12}
		
		\begin{tabularx}{\textwidth}{@{}P{2.0cm} >{\raggedright\arraybackslash}X P{2.2cm}@{}}
			\toprule
			\textbf{Model family} & \textbf{Hyperparameter search space} & \textbf{Configurations per target--representation} \\
			\midrule
			
			Random forest &
			\texttt{n\_estimators} \(\in\{80,120,200,300,400\}\); \newline
			\texttt{max\_depth} \(\in\{\texttt{None},3,4,6,8,12\}\); \newline
			\texttt{min\_samples\_leaf} \(\in\{1,2,3,5,8,12\}\); \newline
			\texttt{min\_samples\_split} \(\in\{2,4,8,12,16\}\); \newline
			\texttt{max\_features} \(\in\{\texttt{sqrt},0.25,0.35,0.50,0.75,1.00\}\); \newline
			\texttt{bootstrap} \(\in\{\texttt{True},\texttt{False}\}\). &
			200 randomized \\
			
			\addlinespace
			
			Extra Trees &
			\texttt{n\_estimators} \(\in\{100,150,250,400,600\}\); \newline
			\texttt{max\_depth} \(\in\{\texttt{None},3,4,6,8,12\}\); \newline
			\texttt{min\_samples\_leaf} \(\in\{1,2,3,5,8,12\}\); \newline
			\texttt{min\_samples\_split} \(\in\{2,4,8,12,16\}\); \newline
			\texttt{max\_features} \(\in\{\texttt{sqrt},0.25,0.35,0.50,0.75,1.00\}\); \newline
			\texttt{bootstrap} \(\in\{\texttt{False},\texttt{True}\}\). &
			200 randomized \\
			
			\addlinespace
			
			Gradient boosting &
			\texttt{n\_estimators} \(\in\{50,80,120,180,250,350\}\); \newline
			\texttt{learning\_rate} \(\in\{0.01,0.02,0.035,0.05,0.08,0.12\}\); \newline
			\texttt{max\_depth} \(\in\{1,2,3\}\); \newline
			\texttt{min\_samples\_leaf} \(\in\{2,3,5,8,12,16\}\); \newline
			\texttt{subsample} \(\in\{0.60,0.75,0.90,1.00\}\); \newline
			\texttt{max\_features} \(\in\{\texttt{None},\texttt{sqrt},0.50,0.80\}\). &
			200 randomized \\
			
			\addlinespace
			
			Histogram gradient boosting &
			\texttt{max\_iter} \(\in\{60,80,120,180,250\}\); \newline
			\texttt{learning\_rate} \(\in\{0.01,0.02,0.035,0.05,0.08,0.12\}\); \newline
			\texttt{max\_leaf\_nodes} \(\in\{5,7,9,15,21,31\}\); \newline
			\texttt{min\_samples\_leaf} \(\in\{5,8,12,16,24,32\}\); \newline
			\texttt{l2\_regularization} \(\in\{0,0.01,0.05,0.10,0.50,1.00,3.00\}\); \newline
			\texttt{early\_stopping} \(\in\{\texttt{True},\texttt{False}\}\). &
			200 randomized \\
			
			\addlinespace
			
			Ridge regression &
			\(\alpha\in\{0.01,0.03,0.10,0.30,1,3,10,30,100,300\}\). &
			10 exhaustive \\
			
			\addlinespace
			
			ElasticNet &
			\(\alpha\in\{0.001,0.003,0.01,0.03,0.10,0.30,1,3,10\}\); \newline
			\texttt{l1\_ratio} \(\in\{0.05,0.10,0.20,0.35,0.50,0.70,0.90\}\); \newline
			\texttt{max\_iter} \(\in\{5000,10000,20000\}\). &
			189 exhaustive \\
			
			\bottomrule
		\end{tabularx}
		
		\vspace{0.2cm}
		
		\begin{minipage}{0.98\textwidth}
			\footnotesize
			\textit{Note.} Each family was tuned separately for four targets and three representations using negative MAE, five recipe-grouped folds, and seed 42.
		\end{minipage}
	\end{table}
	
	\begin{table}[H]
		\centering
		\caption{Hyperparameter search spaces used for inverse dye-recipe reconstruction.}
		\label{tab:inverse_hyperparameter_search}
		\scriptsize
		\setlength{\tabcolsep}{4pt}
		\renewcommand{\arraystretch}{1.12}
		
		\begin{tabularx}{\textwidth}{@{}P{2.0cm} >{\raggedright\arraybackslash}X P{2.2cm}@{}}
			\toprule
			\textbf{Model family} & \textbf{Hyperparameter search space or sampling rule} & \textbf{Configurations per representation} \\
			\midrule
			
			K-nearest neighbours &
			\texttt{n\_neighbors} \(\in\{3,5,7,9,11,15,21\}\); \newline
			\texttt{weights} \(\in\{\texttt{uniform},\texttt{distance}\}\); \newline
			Minkowski parameter \(p\in\{1,2\}\). &
			28 exhaustive \\
			
			\addlinespace
			
			Ridge regression &
			The fixed values \(\alpha\in\{1,3,10\}\) were included. The remaining unique values were sampled as
			\(\alpha=10^{u}\), where \(u\sim\mathcal{U}(-3,2)\). &
			200 sampled \\
			
			\addlinespace
			
			ElasticNet &
			The fixed pairs \((\alpha,l_1\text{-ratio})=(0.01,0.10)\) and \((0.03,0.30)\) were included. Remaining values used
			\(\alpha=10^{u}\), where \(u\sim\mathcal{U}(-4,-0.3)\), and
			\(\texttt{l1\_ratio}\in\{0.05,0.10,0.20,0.40,0.70,0.90\}\).
			The maximum number of iterations was fixed at 20,000. &
			200 sampled \\
			
			\addlinespace
			
			Random forest &
			\texttt{n\_estimators} \(\in\{120,200,300,500\}\); \newline
			\texttt{max\_depth} \(\in\{\texttt{None},4,6,8,12\}\); \newline
			\texttt{min\_samples\_leaf} \(\in\{1,2,3,5,8\}\); \newline
			\texttt{max\_features} \(\in\{\texttt{sqrt},\texttt{log2},0.35,0.50,0.75,1.00\}\).
			Bootstrap sampling was fixed to \texttt{True}. &
			200 sampled \\
			
			\addlinespace
			
			Extra Trees &
			\texttt{n\_estimators} \(\in\{120,200,300,500\}\); \newline
			\texttt{max\_depth} \(\in\{\texttt{None},4,6,8,12\}\); \newline
			\texttt{min\_samples\_leaf} \(\in\{1,2,3,5,8\}\); \newline
			\texttt{max\_features} \(\in\{\texttt{sqrt},\texttt{log2},0.35,0.50,0.75,1.00\}\).
			Bootstrap sampling was fixed to \texttt{False}. &
			200 sampled \\
			
			\bottomrule
		\end{tabularx}
		
		\vspace{0.2cm}
		
		\begin{minipage}{0.98\textwidth}
			\footnotesize
			\textit{Note.} Models were tuned for target-spectral and target-spectral-process inputs using five recipe-grouped folds and seed 42. Selection minimized
			\(S_m=\mathrm{MAE}_{\mathrm{recipe},m}+0.1\,\mathrm{MAE}_{\mathrm{total},m}+(1-F_{1,m}^{\mathrm{active}})\)
			after sparse post-processing. 1,656 configurations were evaluated.
		\end{minipage}
	\end{table}
	
	\clearpage
	
	\begin{landscape}
	\section*{Appendix B: Supplementary Forward and Uncertainty Results}
		
	\subsection*{B1. Extended Forward-Model Leaderboard}
		\label{sec:appendix_forward_leaderboard}
		\setlength{\tabcolsep}{3pt}
		\renewcommand{\arraystretch}{1.15}
		\scriptsize
		\begin{longtable}{
				@{}
				P{2.2cm}
				c
				P{3.6cm}
				P{3.1cm}
				c
				c
				c
				c
				c
				@{}
			}
			\caption{Five lowest pooled out-of-fold MAE models for each color-quality target.}
			\label{tab:extended_forward_leaderboard}\\
			
			\toprule
			\textbf{Target} &
			\textbf{Rank} &
			\textbf{Model} &
			\textbf{Input representation} &
			\textbf{MAE} &
			\textbf{RMSE} &
			\textbf{Median AE} &
			\textbf{Bias} &
			\textbf{Spearman \(\boldsymbol{\rho}\)} \\
			\midrule
			\endfirsthead
			
			\multicolumn{9}{c}
			{{\tablename\ \thetable{} -- continued}}\\
			\toprule
			\textbf{Target} &
			\textbf{Rank} &
			\textbf{Model} &
			\textbf{Input representation} &
			\textbf{MAE} &
			\textbf{RMSE} &
			\textbf{Median AE} &
			\textbf{Bias} &
			\textbf{Spearman \(\boldsymbol{\rho}\)} \\
			\midrule
			\endhead
			
			\midrule
			\multicolumn{9}{r}{Continued on next page}\\
			\endfoot
			
			\bottomrule
			\endlastfoot
			
			\multirow{5}{2.2cm}{\(\Delta E_{\mathrm{D65}}\)}
			& 1
			& Histogram gradient boosting
			& Spectral-only
			& 0.1655
			& 0.2094
			& 0.1558
			& 0.0069
			& 0.2368 \\
			
			& 2
			& Random forest
			& Spectral-only
			& 0.1657
			& 0.2130
			& 0.1359
			& 0.0039
			& 0.2422 \\
			
			& 3
			& Histogram gradient boosting
			& Recipe--process
			& 0.1669
			& 0.2121
			& 0.1578
			& 0.0033
			& 0.2455 \\
			
			& 4
			& Gradient boosting
			& Spectral-only
			& 0.1670
			& 0.2107
			& 0.1448
			& 0.0067
			& 0.2661 \\
			
			& 5
			& Gradient boosting
			& Tree-safe full
			& 0.1670
			& 0.2124
			& 0.1436
			& 0.0073
			& 0.2224 \\
			
			\addlinespace[3pt]
			\midrule
			\addlinespace[3pt]
			
			\multirow{5}{2.2cm}{\(\Delta E_{\mathrm{F11}}\)}
			& 1
			& Gradient boosting
			& Recipe--process
			& 0.1523
			& 0.1977
			& 0.1279
			& 0.0040
			& 0.3988 \\
			
			& 2
			& Gradient boosting
			& Tree-safe full
			& 0.1534
			& 0.1986
			& 0.1251
			& 0.0035
			& 0.3524 \\
			
			& 3
			& Random forest
			& Recipe--process
			& 0.1560
			& 0.2064
			& 0.1180
			& \(-0.0012\)
			& 0.3144 \\
			
			& 4
			& Extra Trees
			& Recipe--process
			& 0.1569
			& 0.2076
			& 0.1262
			& 0.0023
			& 0.2503 \\
			
			& 5
			& Quantile random forest
			& Tree-safe plus K/S
			& 0.1572
			& 0.2073
			& 0.1325
			& \(-0.0104\)
			& 0.2508 \\
			
			\addlinespace[3pt]
			\midrule
			\addlinespace[3pt]
			
			\multirow{5}{2.2cm}{\(\Delta E_{\max}\)}
			& 1
			& Histogram gradient boosting
			& Spectral-only
			& 0.1398
			& 0.1822
			& 0.1228
			& 0.0087
			& 0.2507 \\
			
			& 2
			& Random forest
			& Tree-safe full
			& 0.1423
			& 0.1826
			& 0.1296
			& 0.0035
			& 0.2472 \\
			
			& 3
			& Random forest
			& Spectral-only
			& 0.1433
			& 0.1829
			& 0.1303
			& 0.0021
			& 0.1899 \\
			
			& 4
			& Gradient boosting
			& Spectral-only
			& 0.1434
			& 0.1830
			& 0.1247
			& 0.0042
			& 0.2528 \\
			
			& 5
			& Gradient boosting
			& Tree-safe full
			& 0.1434
			& 0.1851
			& 0.1209
			& 0.0006
			& 0.2321 \\
			
			\addlinespace[3pt]
			\midrule
			\addlinespace[3pt]
			
			\multirow{5}{2.2cm}{Absolute metamerism gap}
			& 1
			& Quantile random forest
			& Linear-safe plus K/S
			& 0.1046
			& 0.1364
			& 0.0865
			& \(-0.0083\)
			& 0.4746 \\
			
			& 2
			& Random forest
			& Tree-safe full
			& 0.1054
			& 0.1394
			& 0.0803
			& 0.0014
			& 0.4161 \\
			
			& 3
			& Gradient boosting
			& Tree-safe full
			& 0.1077
			& 0.1374
			& 0.0941
			& \(-0.0092\)
			& 0.4992 \\
			
			& 4
			& Quantile random forest
			& Tree-safe plus K/S
			& 0.1083
			& 0.1457
			& 0.0750
			& \(-0.0062\)
			& 0.3619 \\
			
			& 5
			& Random forest
			& Spectral-only
			& 0.1083
			& 0.1433
			& 0.0894
			& \(-0.0071\)
			& 0.4447 \\
			
		\end{longtable}
		
		\vspace{-0.3cm}
		
		\begin{minipage}{\linewidth}
			\footnotesize
			\textit{Note.} Metrics were calculated from pooled out-of-fold predictions under five-fold recipe-grouped cross-validation.
		\end{minipage}
		\end{landscape}
	
	\begin{landscape}
	\subsection*{B2. Complete Uncertainty-Method Comparison}
	\label{sec:appendix_uncertainty_comparison}
		\scriptsize
		\setlength{\tabcolsep}{2pt}
		\renewcommand{\arraystretch}{1.15}

		\begin{longtable}{
				@{}
				P{1.5cm}
				P{1.9cm}
				P{2.0cm}
				c
				P{1.75cm}
				c
				P{1.75cm}
				c
				c
				c
				c
				c
				@{}
			}
			\caption{Complete comparison of uncertainty methods for the four color-quality targets.}
			\label{tab:complete_uncertainty_comparison}\\
			
			\toprule
			\textbf{Target} &
			\textbf{Method} &
			\textbf{Input representation} &
			\shortstack{\textbf{Median-}\\\textbf{prediction}\\\textbf{MAE}} &
			\textbf{95\% CI for MAE} &
			\textbf{Coverage} &
			\textbf{95\% CI for coverage} &
			\shortstack{\textbf{Mean}\\\textbf{width}} &
			\shortstack{\textbf{Pinball}\\\textbf{loss}} &
			\shortstack{\textbf{Width--error}\\\textbf{Spearman \(\boldsymbol{\rho}\)}} &
			\shortstack{\textbf{Coverage}\\\textbf{compatible}} &
			\textbf{Selected} \\
			\midrule
			\endfirsthead
			
			\multicolumn{12}{c}
			{{\tablename\ \thetable{} -- continued}}\\
			
			\toprule
			\textbf{Target} &
			\textbf{Method} &
			\textbf{Input representation} &
			\shortstack{\textbf{Median-}\\\textbf{prediction}\\\textbf{MAE}} &
			\textbf{95\% CI for MAE} &
			\textbf{Coverage} &
			\textbf{95\% CI for coverage} &
			\shortstack{\textbf{Mean}\\\textbf{width}} &
			\shortstack{\textbf{Pinball}\\\textbf{loss}} &
			\shortstack{\textbf{Width--error}\\\textbf{Spearman \(\boldsymbol{\rho}\)}} &
			\shortstack{\textbf{Coverage}\\\textbf{compatible}} &
			\textbf{Selected} \\
			\midrule
			\endhead
			
			\midrule
			\multicolumn{12}{r}{Continued on next page}\\
			\endfoot
			
			\bottomrule
			\endlastfoot
			
			\multirow{3}{1.5cm}{\(\Delta E_{\mathrm{D65}}\)}
			&
			Fold-wise empirical quantiles
			&
			Response only
			&
			0.1733
			&
			0.1546--0.1933
			&
			0.7877
			&
			0.7277--0.8436
			&
			0.5520
			&
			0.0517
			&
			0.0512
			&
			Yes
			&
			Yes
			\\
			
			&
			Quantile random forest
			&
			Linear-safe plus K/S
			&
			0.1762
			&
			0.1585--0.1960
			&
			0.4387
			&
			0.3689--0.5139
			&
			0.2836
			&
			0.0611
			&
			\(-0.0228\)
			&
			No
			&
			No
			\\
			
			&
			Quantile random forest
			&
			Tree-safe plus K/S
			&
			0.1749
			&
			0.1559--0.1962
			&
			0.4670
			&
			0.3981--0.5415
			&
			0.3043
			&
			0.0599
			&
			\(-0.0322\)
			&
			No
			&
			No
			\\
			
			\addlinespace[3pt]
			\midrule
			\addlinespace[3pt]
			
			\multirow{3}{1.5cm}{\(\Delta E_{\mathrm{F11}}\)}
			&
			Fold-wise empirical quantiles
			&
			Response only
			&
			0.1676
			&
			0.1451--0.1925
			&
			0.7925
			&
			0.7297--0.8512
			&
			0.4971
			&
			0.0522
			&
			\(-0.0600\)
			&
			Yes
			&
			Yes
			\\
			
			&
			Quantile random forest
			&
			Linear-safe plus K/S
			&
			0.1599
			&
			0.1361--0.1859
			&
			0.5283
			&
			0.4465--0.6073
			&
			0.2580
			&
			0.0571
			&
			0.0386
			&
			No
			&
			No
			\\
			
			&
			Quantile random forest
			&
			Tree-safe plus K/S
			&
			0.1572
			&
			0.1366--0.1797
			&
			0.4151
			&
			0.3333--0.4951
			&
			0.2124
			&
			0.0589
			&
			0.1300
			&
			No
			&
			No
			\\
			
			\addlinespace[3pt]
			\midrule
			\addlinespace[3pt]
			
			\multirow{3}{1.5cm}{\(\Delta E_{\max}\)}
			&
			Fold-wise empirical quantiles
			&
			Response only
			&
			0.1485
			&
			0.1311--0.1687
			&
			0.7972
			&
			0.7281--0.8585
			&
			0.4826
			&
			0.0455
			&
			\(-0.0678\)
			&
			Yes
			&
			Yes
			\\
			
			&
			Quantile random forest
			&
			Linear-safe plus K/S
			&
			0.1458
			&
			0.1291--0.1659
			&
			0.2783
			&
			0.2123--0.3427
			&
			0.1401
			&
			0.0582
			&
			0.0178
			&
			No
			&
			No
			\\
			
			&
			Quantile random forest
			&
			Tree-safe plus K/S
			&
			0.1449
			&
			0.1286--0.1644
			&
			0.2736
			&
			0.2120--0.3395
			&
			0.1271
			&
			0.0590
			&
			\(-0.0118\)
			&
			No
			&
			No
			\\
			
			\addlinespace[3pt]
			\midrule
			\addlinespace[3pt]
			
			\multirow{3}{1.5cm}{Absolute metamerism gap}
			&
			Fold-wise empirical quantiles
			&
			Response only
			&
			0.1319
			&
			0.1162--0.1477
			&
			0.8066
			&
			0.7500--0.8616
			&
			0.4104
			&
			0.0387
			&
			\(-0.0414\)
			&
			Yes
			&
			Yes
			\\
			
			&
			Quantile random forest
			&
			Linear-safe plus K/S
			&
			0.1046
			&
			0.0906--0.1193
			&
			0.7406
			&
			0.6732--0.8088
			&
			0.3240
			&
			0.0326
			&
			0.2999
			&
			Yes
			&
			No
			\\
			
			&
			Quantile random forest
			&
			Tree-safe plus K/S
			&
			0.1083
			&
			0.0918--0.1255
			&
			0.6179
			&
			0.5395--0.6952
			&
			0.2715
			&
			0.0354
			&
			0.3443
			&
			No
			&
			No
			\\
			
		\end{longtable}
		
		\vspace{-0.3cm}
		
		\begin{minipage}{\linewidth}
			\footnotesize
			\textit{Note.} Nominal coverage was 0.80. Coverage was considered compatible when its 95\% cluster-bootstrap confidence interval included 0.80.
		\end{minipage}
	\end{landscape}
	
	\begin{landscape}
	\section*{Appendix C: Supplementary Inverse Results}
	\subsection*{C1. Complete Inverse Base-Model Comparison}
	\label{sec:appendix_inverse_base_models}
	
		\begin{table}[htbp]
			\centering
			\caption{Complete comparison of tuned inverse base models under recipe-grouped cross-validation.}
			\label{tab:complete_inverse_base_models}
			\scriptsize
			\setlength{\tabcolsep}{3pt}
			\renewcommand{\arraystretch}{1.15}
			
			\begin{adjustbox}{max width=\linewidth}
				\begin{tabular}{
						@{}
						c
						P{2.4cm}
						P{3.1cm}
						c
						c
						c
						c
						c
						c
						c
						@{}
					}
					\toprule
					\textbf{Rank} &
					\textbf{Model} &
					\textbf{Input representation} &
					\shortstack{\textbf{Recipe-vector}\\\textbf{MAE}} &
					\shortstack{\textbf{Total-dye}\\\textbf{absolute error}} &
					\shortstack{\textbf{Active-dye}\\\textbf{precision}} &
					\shortstack{\textbf{Active-dye}\\\textbf{recall}} &
					\shortstack{\textbf{Active-dye}\\\textbf{F1}} &
					\shortstack{\textbf{Active Top-\(k\)}\\\textbf{recall}} &
					\shortstack{\textbf{Composite}\\\textbf{score}} \\
					\midrule
					
					1 &
					Extra Trees &
					Target spectral &
					0.0338 &
					0.3223 &
					0.6824 &
					0.6942 &
					0.6855 &
					0.6892 &
					0.3805 \\
					
					2 &
					Extra Trees &
					Target spectral and process &
					0.0319 &
					0.2723 &
					0.6682 &
					0.6965 &
					0.6777 &
					0.6989 &
					0.3814 \\
					
					3 &
					Random forest &
					Target spectral &
					0.0334 &
					0.3237 &
					0.6745 &
					0.6981 &
					0.6821 &
					0.6787 &
					0.3837 \\
					
					4 &
					KNN &
					Target spectral &
					0.0327 &
					0.2985 &
					0.6722 &
					0.6816 &
					0.6736 &
					0.6908 &
					0.3889 \\
					
					5 &
					Random forest &
					Target spectral and process &
					0.0341 &
					0.2862 &
					0.6612 &
					0.6895 &
					0.6711 &
					0.6626 &
					0.3916 \\
					
					6 &
					KNN &
					Target spectral and process &
					0.0310 &
					0.3108 &
					0.6675 &
					0.6769 &
					0.6682 &
					0.6932 &
					0.3938 \\
					
					7 &
					ElasticNet &
					Target spectral &
					0.0526 &
					0.6330 &
					0.4717 &
					0.4741 &
					0.4726 &
					0.4855 &
					0.6433 \\
					
					8 &
					Ridge &
					Target spectral &
					0.0549 &
					0.6876 &
					0.4654 &
					0.4678 &
					0.4664 &
					0.4791 &
					0.6573 \\
					
					9 &
					ElasticNet &
					Target spectral and process &
					0.0614 &
					0.4468 &
					0.3947 &
					0.3947 &
					0.3947 &
					0.4042 &
					0.7114 \\
					
					10 &
					Ridge &
					Target spectral and process &
					0.0649 &
					0.6138 &
					0.4057 &
					0.4080 &
					0.4066 &
					0.4179 &
					0.7196 \\
					
					\bottomrule
				\end{tabular}
			\end{adjustbox}
			
			\vspace{0.15cm}
			
			\begin{minipage}{0.96\linewidth}
				\footnotesize
				\textit{Note.} Metrics use sparse out-of-fold predictions; concentration errors are in percentage points (\% OWF). Ranks follow the unrounded composite score, with lower values indicating better performance.
			\end{minipage}
		\end{table}
	\end{landscape}

	\begin{landscape}
	\subsection*{C2. Meta-model comparison}
	\label{sec:appendix_inverse_meta_models}

		\begin{table}[H]
			\centering
			\caption{Comparison of inverse meta-model strategies under recipe-grouped cross-validation.}
			\label{tab:inverse_meta_model_comparison}
			\scriptsize
			\setlength{\tabcolsep}{3.5pt}
			\renewcommand{\arraystretch}{1.15}
			
			\begin{adjustbox}{max width=\linewidth}
				\begin{tabular}{
						@{}
						c
						P{4.0cm}
						c
						c
						c
						c
						c
						c
						c
						@{}
					}
					\toprule
					\textbf{Rank} &
					\textbf{Meta-model} &
					\shortstack{\textbf{Recipe-vector}\\\textbf{MAE}} &
					\shortstack{\textbf{Total-dye}\\\textbf{absolute error}} &
					\shortstack{\textbf{Active-dye}\\\textbf{precision}} &
					\shortstack{\textbf{Active-dye}\\\textbf{recall}} &
					\shortstack{\textbf{Active-dye}\\\textbf{F1}} &
					\shortstack{\textbf{Active Top-\(k\)}\\\textbf{recall}} &
					\shortstack{\textbf{Composite}\\\textbf{score}} \\
					\midrule
					
					1 &
					Component-wise median blend &
					0.0295 &
					0.3297 &
					0.7075 &
					0.6808 &
					0.6887 &
					0.6932 &
					0.3737 \\
					
					2 &
					ElasticNet meta-model &
					0.0382 &
					0.4809 &
					0.5519 &
					0.5715 &
					0.5583 &
					0.5612 &
					0.5280 \\
					
					3 &
					Arithmetic-mean blend &
					0.0402 &
					0.3582 &
					0.5252 &
					0.5401 &
					0.5297 &
					0.5362 &
					0.5463 \\
					
					4 &
					Ridge meta-model &
					0.0399 &
					0.5124 &
					0.5362 &
					0.5432 &
					0.5390 &
					0.5491 &
					0.5521 \\
					
					\bottomrule
				\end{tabular}
			\end{adjustbox}
			
			\vspace{0.15cm}
			
			\begin{minipage}{0.96\linewidth}
				\footnotesize
				\textit{Note.} Metrics use sparse predictions derived from ten base-model outputs; concentration errors are in percentage points (\% OWF). Ranks follow the unrounded composite score, with lower values indicating better performance.
			\end{minipage}
		\end{table}
		
	\end{landscape}
	
	\clearpage
	\begin{landscape}
	\subsection*{C3. Strict Unseen-Target Comparison}
	\label{sec:appendix_strict_inverse_comparison}
	
	\begin{table}[htbp]
			\centering
			\caption{Comparison of inverse base models under target-grouped cross-validation.}
			\label{tab:strict_inverse_model_comparison}
			\scriptsize
			\setlength{\tabcolsep}{3pt}
			\renewcommand{\arraystretch}{1.15}
			
			\begin{adjustbox}{max width=\linewidth}
				\begin{tabular}{
						@{}
						c
						P{2.5cm}
						P{3.2cm}
						c
						c
						c
						c
						c
						c
						c
						@{}
					}
					\toprule
					\textbf{Rank} &
					\textbf{Model} &
					\textbf{Input representation} &
					\shortstack{\textbf{Recipe-vector}\\\textbf{MAE}} &
					\shortstack{\textbf{Total-dye}\\\textbf{absolute error}} &
					\shortstack{\textbf{Active-dye}\\\textbf{precision}} &
					\shortstack{\textbf{Active-dye}\\\textbf{recall}} &
					\shortstack{\textbf{Active-dye}\\\textbf{F1}} &
					\shortstack{\textbf{Active Top-\(k\)}\\\textbf{recall}} &
					\shortstack{\textbf{Composite}\\\textbf{score}} \\
					\midrule
					
					1 &
					Random forest &
					Target spectral and process &
					0.0602 &
					0.4556 &
					0.3931 &
					0.4025 &
					0.3969 &
					0.4122 &
					0.7089 \\
					
					2 &
					Extra Trees &
					Target spectral &
					0.0629 &
					0.4900 &
					0.3954 &
					0.4096 &
					0.4009 &
					0.4147 &
					0.7109 \\
					
					3 &
					Random forest &
					Target spectral &
					0.0613 &
					0.4703 &
					0.3664 &
					0.3758 &
					0.3701 &
					0.3849 &
					0.7382 \\
					
					4 &
					Extra Trees &
					Target spectral and process &
					0.0652 &
					0.4387 &
					0.3522 &
					0.3616 &
					0.3560 &
					0.3704 &
					0.7531 \\
					
					5 &
					KNN &
					Target spectral and process &
					0.0633 &
					0.4344 &
					0.3318 &
					0.3349 &
					0.3330 &
					0.3430 &
					0.7738 \\
					
					6 &
					KNN &
					Target spectral &
					0.0717 &
					0.6116 &
					0.3082 &
					0.3153 &
					0.3110 &
					0.3156 &
					0.8219 \\
					
					7 &
					ElasticNet &
					Target spectral and process &
					0.0771 &
					0.5453 &
					0.3082 &
					0.3082 &
					0.3082 &
					0.3156 &
					0.8235 \\
					
					8 &
					Ridge &
					Target spectral and process &
					0.0857 &
					0.7711 &
					0.3035 &
					0.3058 &
					0.3044 &
					0.3060 &
					0.8584 \\
					
					9 &
					ElasticNet &
					Target spectral &
					0.0938 &
					0.9794 &
					0.2657 &
					0.2681 &
					0.2667 &
					0.2746 &
					0.9250 \\
					
					10 &
					Ridge &
					Target spectral &
					0.0984 &
					1.1102 &
					0.2720 &
					0.2744 &
					0.2730 &
					0.2810 &
					0.9364 \\
					
					\bottomrule
				\end{tabular}
			\end{adjustbox}
			
			\vspace{0.15cm}
		\end{table}
	\end{landscape}

	\clearpage
	\subsection*{C4. Full Dye-Specific Results}
	\label{sec:appendix_dye_specific_results}
	
		
		\setlength{\tabcolsep}{4pt}
		\renewcommand{\arraystretch}{1.12}
		
		\begin{longtable}{
				@{}
				l
				r
				r
				r
				r
				r
				r
				r
				r
				@{}
			}

			\caption{Dye-specific reconstruction results for the selected
				target-spectral Extra Trees model.}
			\label{tab:full_dye_specific_results}\\
			\toprule
			\textbf{Dye} &
			\shortstack{\textbf{Observed}\\\textbf{active}} &
			\shortstack{\textbf{Predicted}\\\textbf{active}} &
			\textbf{MAE} &
			\textbf{RMSE} &
			\textbf{Bias} &
			\textbf{Precision} &
			\textbf{Recall} &
			\textbf{F1} \\
			\midrule
			\endfirsthead
			
			\multicolumn{9}{c}
			{{\tablename\ \thetable{} -- continued}}\\
			
			\toprule
			\textbf{Dye} &
			\shortstack{\textbf{Observed}\\\textbf{active}} &
			\shortstack{\textbf{Predicted}\\\textbf{active}} &
			\textbf{MAE} &
			\textbf{RMSE} &
			\textbf{Bias} &
			\textbf{Precision} &
			\textbf{Recall} &
			\textbf{F1} \\
			\midrule
			\endhead
			
			\midrule
			\multicolumn{9}{r}{Continued on next page}\\
			\endfoot
			
			\bottomrule
			\endlastfoot
			
			Dye A  & 65 & 69 & 0.12570 & 0.29253 &  0.01670 & 0.8406 & 0.8923 & 0.8657 \\
			Dye B  & 70 & 87 & 0.24896 & 0.63326 &  0.10403 & 0.6782 & 0.8429 & 0.7516 \\
			Dye C  & 42 & 39 & 0.02160 & 0.08548 & -0.00903 & 1.0000 & 0.9286 & 0.9630 \\
			Dye D  &  0 &  0 & 0.00000 & 0.00000 &  0.00000 & 0.0000 & 0.0000 & 0.0000 \\
			Dye E  & 54 & 49 & 0.10598 & 0.28474 & -0.01246 & 0.7551 & 0.6852 & 0.7184 \\
			Dye F  & 31 & 30 & 0.11577 & 0.43177 & -0.00616 & 0.6667 & 0.6452 & 0.6557 \\
			Dye G  &  9 & 14 & 0.03655 & 0.19221 &  0.01574 & 0.6429 & 1.0000 & 0.7826 \\
			Dye H  & 12 & 14 & 0.02664 & 0.10431 &  0.00175 & 0.4286 & 0.5000 & 0.4615 \\
			Dye I  & 42 & 45 & 0.06863 & 0.20318 & -0.01938 & 0.6000 & 0.6429 & 0.6207 \\
			Dye J  & 40 & 50 & 0.04685 & 0.13189 &  0.01384 & 0.7000 & 0.8750 & 0.7778 \\
			Dye K  & 41 & 43 & 0.02476 & 0.08492 &  0.00835 & 0.7674 & 0.8049 & 0.7857 \\
			Dye L  &  9 &  6 & 0.09076 & 0.36750 & -0.01598 & 0.0000 & 0.0000 & 0.0000 \\
			Dye M  & 33 & 33 & 0.04176 & 0.15300 & -0.02328 & 0.6364 & 0.6364 & 0.6364 \\
			Dye N  & 19 & 29 & 0.02674 & 0.10712 &  0.01957 & 0.6552 & 1.0000 & 0.7917 \\
			Dye O  & 14 &  7 & 0.03207 & 0.13581 & -0.01476 & 0.5714 & 0.2857 & 0.3810 \\
			Dye P  &  3 &  3 & 0.04567 & 0.27572 & -0.00810 & 0.0000 & 0.0000 & 0.0000 \\
			Dye Q  &  8 & 24 & 0.04673 & 0.15608 &  0.01912 & 0.2500 & 0.7500 & 0.3750 \\
			Dye R  &  4 &  6 & 0.02654 & 0.14180 & -0.00610 & 0.0000 & 0.0000 & 0.0000 \\
			Dye S  &  4 &  3 & 0.01075 & 0.07575 & -0.00981 & 0.0000 & 0.0000 & 0.0000 \\
			Dye T  &  3 &  3 & 0.02061 & 0.12605 & -0.00501 & 0.0000 & 0.0000 & 0.0000 \\
			Dye U  & 15 & 11 & 0.01208 & 0.07272 & -0.00626 & 0.9091 & 0.6667 & 0.7692 \\
			Dye V  & 13 & 13 & 0.00396 & 0.01929 & -0.00054 & 0.7692 & 0.7692 & 0.7692 \\
			Dye W  &  2 &  0 & 0.01028 & 0.10591 & -0.01028 & 0.0000 & 0.0000 & 0.0000 \\
			Dye X  &  1 &  0 & 0.00830 & 0.12088 & -0.00830 & 0.0000 & 0.0000 & 0.0000 \\
			Dye Y  & 15 &  7 & 0.00386 & 0.02546 & -0.00375 & 1.0000 & 0.4667 & 0.6364 \\
			Dye Z  &  1 &  0 & 0.00448 & 0.06525 & -0.00448 & 0.0000 & 0.0000 & 0.0000 \\
			Dye AA & 13 & 19 & 0.00163 & 0.00683 &  0.00163 & 0.6842 & 1.0000 & 0.8125 \\
			Dye AB &  1 &  0 & 0.00208 & 0.03022 & -0.00208 & 0.0000 & 0.0000 & 0.0000 \\
			Dye AC &  2 &  2 & 0.00257 & 0.02169 & -0.00150 & 0.0000 & 0.0000 & 0.0000 \\
			Dye AD &  0 &  0 & 0.00000 & 0.00000 &  0.00000 & 0.0000 & 0.0000 & 0.0000 \\
			Dye AE &  2 &  0 & 0.00152 & 0.01569 & -0.00152 & 0.0000 & 0.0000 & 0.0000 \\
			Dye AF &  1 &  0 & 0.00125 & 0.01820 & -0.00125 & 0.0000 & 0.0000 & 0.0000 \\
			Dye AG &  2 &  0 & 0.00119 & 0.01224 & -0.00119 & 0.0000 & 0.0000 & 0.0000 \\
			Dye AH &  3 &  0 & 0.00030 & 0.00203 & -0.00030 & 0.0000 & 0.0000 & 0.0000 \\
			Dye AI &  3 &  0 & 0.00021 & 0.00176 & -0.00021 & 0.0000 & 0.0000 & 0.0000 \\
			Dye AJ &  0 &  0 & 0.00001 & 0.00008 & -0.00001 & 0.0000 & 0.0000 & 0.0000 \\
			
		\end{longtable}
		
		\vspace{-0.3cm}
		
		\begin{minipage}{\linewidth}
			\footnotesize
			\textit{Note.} Active counts use a 0.01\% OWF threshold; MAE, RMSE, and bias are in percentage points (\% OWF). Classification metrics with an undefined denominator were recorded as zero.
		\end{minipage}

	\begin{figure}[H]
		\centering
		\includegraphics[width=0.90\textwidth]{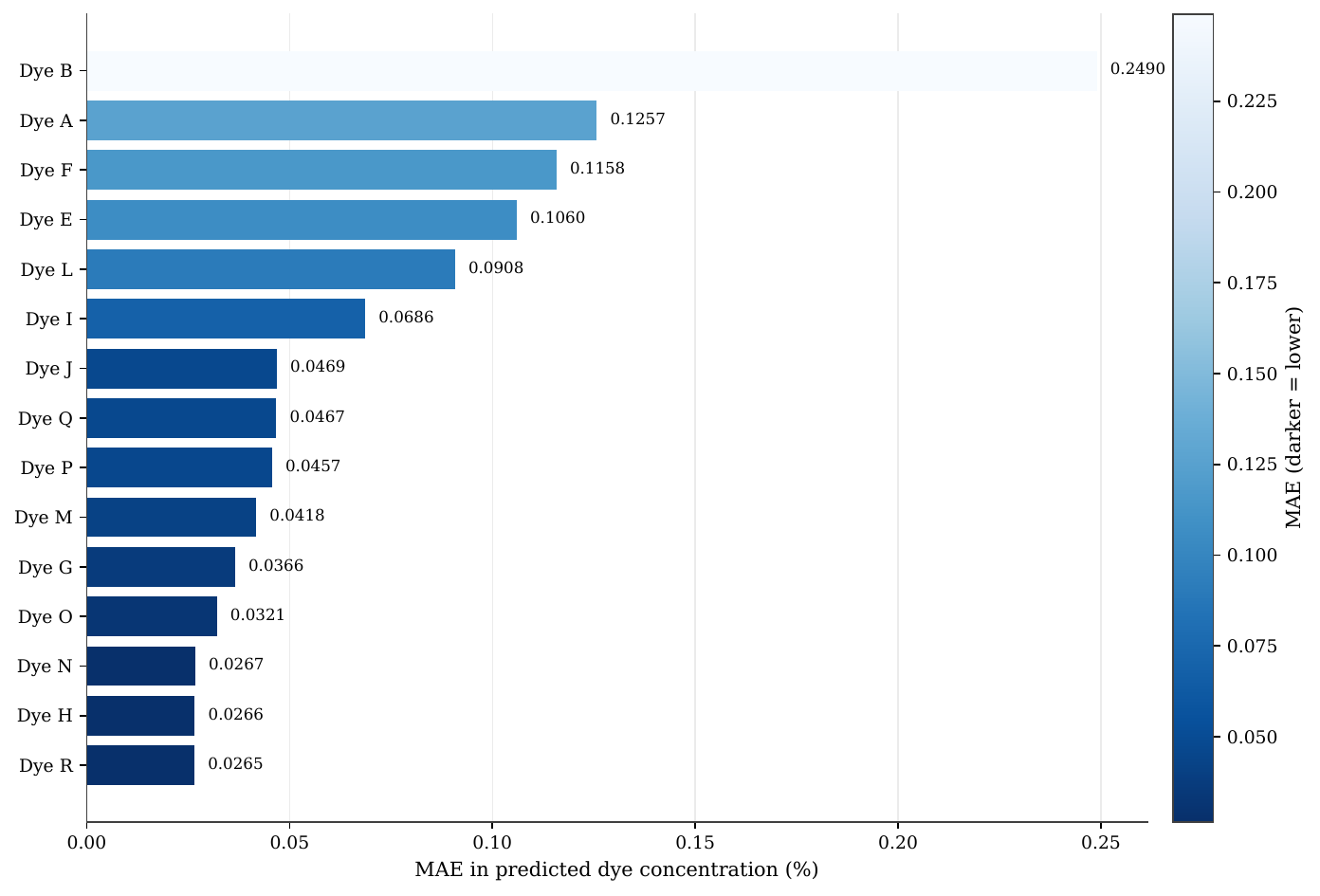}
		\caption{Dye-specific concentration MAE for the 15 most difficult components under the selected grouped-validation inverse model.}
		\label{fig:inverse_dye_specific_error}
	\end{figure}
	
	\subsection*{C5. Supplementary Inverse Figures}
	\label{sec:appendix_inverse_figures}
	
	\begin{figure}[H]
		\centering
		\includegraphics[width=0.8\textwidth]
		{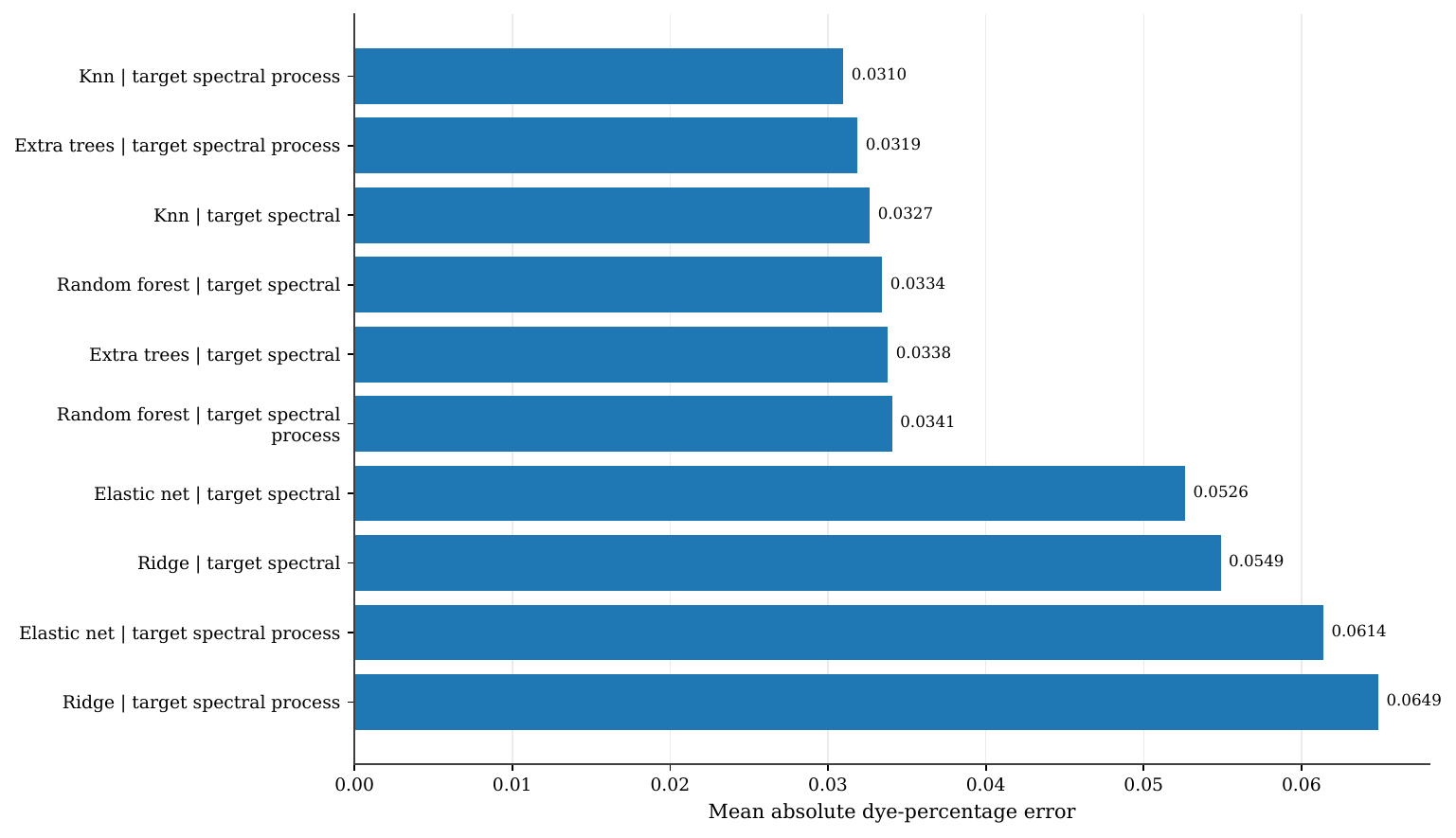}
		\caption{Recipe-vector mean absolute error for the inverse base-model configurations under recipe-grouped cross-validation.}
		\label{fig:appendix_inverse_recipe_mae}
	\end{figure}
	
	\begin{figure}[H]
		\centering
		\includegraphics[width=0.8\textwidth]
		{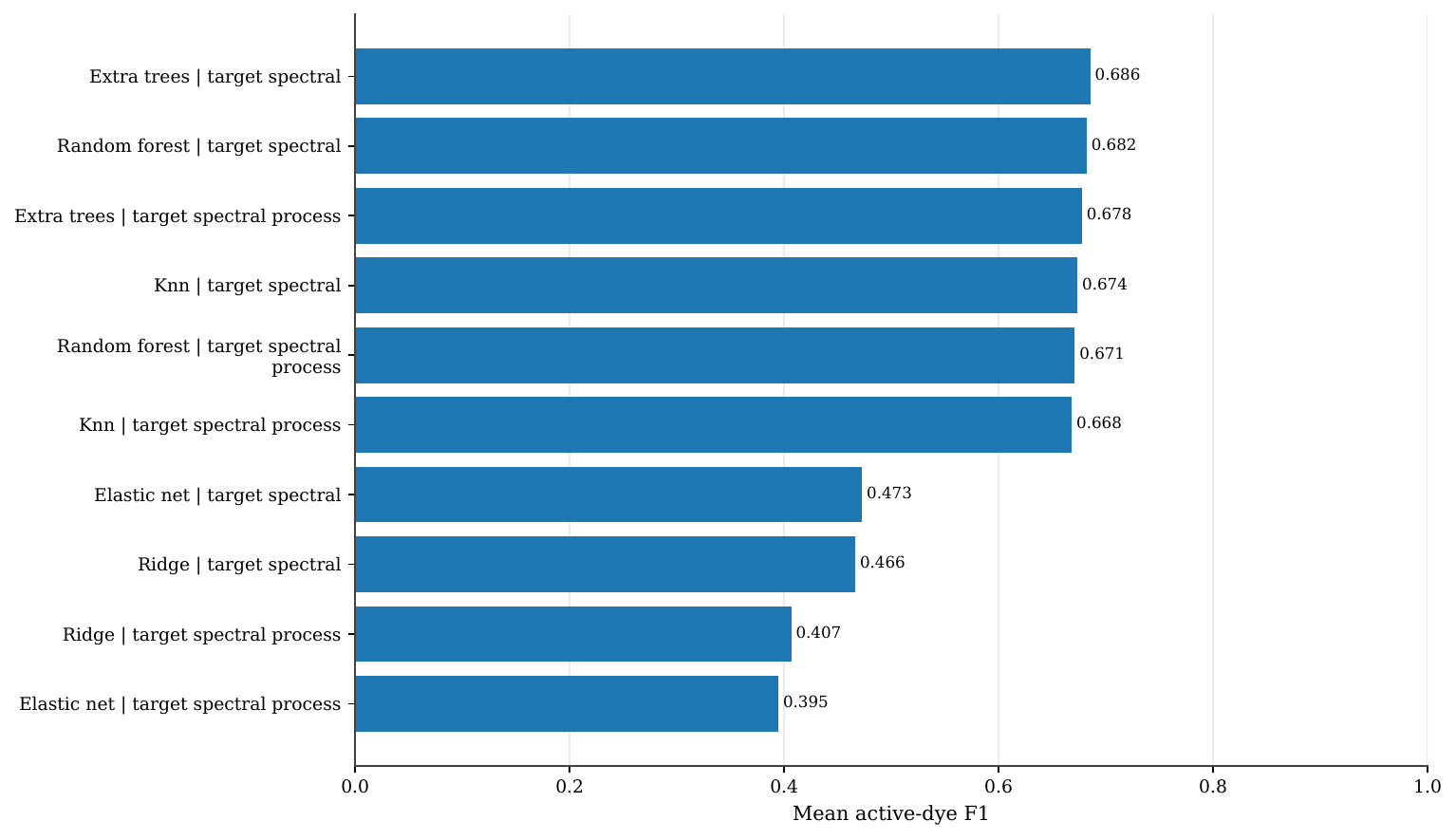}
		\caption{Active-dye F1 scores for the inverse base-model configurations under recipe-grouped cross-validation.}
		\label{fig:appendix_inverse_active_f1}
	\end{figure}
	
	\clearpage
	
	\begin{landscape}
		\section*{Appendix D: Literature Comparison}
		\scriptsize
		\setlength{\tabcolsep}{3pt}
		\renewcommand{\arraystretch}{1.2}
		
		\begin{longtable}{P{2.3cm} P{3.8cm} P{4.6cm} P{3.5cm} P{2.6cm} P{1.2cm}}
			\caption{Different machine-learning architectures and optimization techniques to solve the non-linear dye recipe formulation and color-matching problem} \\
			\hline
			\textbf{Author (Year)} & \textbf{Process} & \textbf{Variables} & \textbf{Dataset} & \textbf{Dataset Size} & \textbf{Country} \\
			\hline
			\endfirsthead
			
			\hline
			\textbf{Author (Year)} & \textbf{Process} & \textbf{Variables} & \textbf{Dataset} & \textbf{Dataset Size} & \textbf{Country} \\
			\hline
			\endhead
			
			\hline
			\endfoot
			
			(Zhu et al., 2025) &
			Full-color-gamut rotor spinning \& Neural Network color prediction &
			Blending ratios (8 primary fibers), spectral reflectance, CIELAB &
			Blended yarn fabrics (experimental samples) &
			273 samples (255 training, 18 validation) &
			China \\
			
			(Cui et al., 2025) &
			Bidirectional color prediction using BPNN models \& Grid-based color mixing &
			Inputs/Outputs: Spectral reflectance, Blend ratios (4 primary colors), CIELAB values &
			Experimental color-spun yarn samples &
			66 distinct yarn samples &
			China \\
			
			(Xie et al., 2024) &
			GBRT (Gradient-Boosting Regression Tree) with Log-transformation &
			Dye concentrations (3 disperse dyes), full reflectance spectra, CIELAB/RGB &
			Industrial dyeing records (Polyester fabrics) &
			356 industrial dyeing data &
			China \\
			
			(Souissi et al., 2024) &
			Genetic Algorithm (GA) optimization (Roulette wheel selection) &
			Disperse dye concentrations (3 dyes), CMC color difference &
			Dyed PET/PTT bicomponent filaments &
			15 reference colors &
			Tunisia \\
			
			(Zhou et al., 2024) &
			NSGA-II (Genetic Algorithm) + ESM-EWM-TOPSIS decision model &
			Dye concentrations, CMC $\Delta E$, metamerism index, cost &
			Laboratory dyeing dataset (Cotton/Reactive dyes) &
			40 target colors &
			China \\
			
			(Sun et al., 2023) &
			Neuronal Network (BPNN) + Gridded color-mixing model &
			Inputs/Outputs: Fiber mixing ratios (4 colors) and Spectral reflectance &
			Gridded color-mixed yarn samples &
			41 samples (31 Training, 10 Validation) &
			China \\
			
			(Vadood \& Haji, 2022) &
			Hybrid ANN-GA (Genetic Algorithm optimized Neural Network) &
			Dyeing parameters (Concentration, Temp, Time, pH), L$^\ast$,a$^\ast$,b$^\ast$ coordinates. &
			Experimental samples (Polyester dyed with Madder natural dye) &
			81 samples &
			Iran \\
			
			(Yang et al., 2017) &
			MANN (Modular Artificial Neural Network) &
			Inputs: Reflectance spectra. Outputs: Fiber blending percentages (Recipe). &
			Top-dyed m\'{e}lange yarn samples &
			500 training samples and 100 blind testing samples &
			China \\
			
			(Chaouch et al., 2020) &
			Genetic Algorithm (GA) optimization &
			Inputs: Dye selection and proportions (3 reactive, 4 direct dyes). Outputs: CMC color difference. &
			Dyed 100\% cotton woven fabrics &
			20 three dye \& 20 four dye samples &
			Tunisia \\
			
			(Li et al., 2022) &
			FWSVR (Feature-Weighted SVR) + PSO (Particle Swarm Optimization) &
			Dye concentrations $\rightarrow$ CIELAB; target color $\rightarrow$ recipe &
			Cotton \& taffeta fabrics &
			Experimental datasets. 200 samples (180 training, 20 validation) &
			China \\
			
			(Zhang et al., 2021) &
			Improved RNN (Recurrent Neural Network) + Hyperspectral measurement &
			Input: Spectral reflectance (3D hyperspectral data). Output: Dyeing concentrations (Recipe). &
			Evenly \& unevenly dyed cotton fabrics &
			363 samples (343 evenly dyed, 20 unevenly dyed) &
			China \\
			
			(Yu et al., 2021) &
			PSO-LSSVM (Least Squares Support Vector Machine optimized by Particle Swarm Optimization) &
			L*, a*, b* $\rightarrow$ dye concentration (single dye) &
			Cotton knitted fabrics (Levafix reactive dyes) &
			75 samples (25 for each of the 3 single-dye databases) &
			China \\
			
			(Qin \& Zhang, 2021) &
			Modular GBRT (Gradient Boosting Regression Tree) &
			Inputs: Target color/fabric type. Outputs: Dye concentrations (DCS). &
			Historical industrial dyeing records (per fabric type + dye combo) &
			810 records &
			China \\
			
			(Safi \& Amirshahi, 2023) &
			Comparison of Kubelka--Munk vs. Allen--Goldfinger reflective models &
			Input: Spectral reflectance. Output: Dye concentration. &
			Nylon 6 fabrics dyed with 4 acid dyes &
			22 concentration levels (0--6\% owf) &
			Iran \\
			
			(\c{S}ahin \& Ey\"{u}po\u{g}lu, 2024) &
			LSTM (Long Short-Term Memory) + Hyperspectral vs. Spectrophotometer comparison &
			Inputs: Spectral reflectance (from hyperspectral camera). Outputs: Target color/recipe estimation. &
			Industrial laboratory samples (Textile finishing company) &
			Not stated &
			Turkey \\
			
		\end{longtable}
		
	\end{landscape}
	
	\clearpage

\end{document}